\documentclass{article}
\PassOptionsToPackage{numbers, compress}{natbib}
\usepackage[final,main]{neurips_2026}

\usepackage[utf8]{inputenc} 
\usepackage[T1]{fontenc}    
\usepackage{hyperref}       
\usepackage{url}            
\usepackage{booktabs}       
\usepackage{amsfonts}       
\usepackage{amsmath, amssymb, amsthm, mathtools} 
\usepackage{graphicx}       
\usepackage[linesnumbered,ruled,vlined]{algorithm2e}
\usepackage{bbm}            
\usepackage{enumitem}
\usepackage{nicefrac}       
\usepackage{microtype}      
\usepackage{xcolor}         
\usepackage{mathrsfs}       
\usepackage{caption}        
\usepackage{float}          
\usepackage{flafter}
\usepackage{placeins}

\DeclareMathOperator*{\argmin}{arg\,min}
\DeclareMathOperator{\Var}{Var}
\DeclareMathOperator{\E}{\mathbb{E}}

\newcommand{\Otilde}{\tilde{\mathcal{O}}}

\newtheorem{theorem}{Theorem}[section]
\newtheorem{lemma}[theorem]{Lemma}
\newtheorem{proposition}[theorem]{Proposition}
\newtheorem{corollary}[theorem]{Corollary}
\newtheorem{definition}[theorem]{Definition}
\newtheorem{remark}[theorem]{Remark}
\newtheorem{assumption}{Assumption}

\title{OPSRL-SSP: Optimistic Posterior Sampling for Stochastic Shortest Path with Minimax-Optimal Regret}

\author{%
  Yiming Xu \quad Xian Wei \quad Cheng Chen\thanks{Corresponding author} \\
  MoE Engineering Research Center of HW/SW
  Co-design Technology and Application\\
  East China Normal University \\
  \texttt{51285902185@stu.ecnu.edu.cn} \quad
  \texttt{\{xwei,chchen\}@sei.ecnu.edu.cn} \\
}

\begin{document}

\maketitle

\begin{abstract}
We study online reinforcement learning for the stochastic shortest path (SSP) problem with unknown transition dynamics. Although existing posterior-sampling methods perform well empirically on SSPs, their regret guarantees do not achieve the minimax-optimal order. We address this limitation by proposing OPSRL-SSP, an optimistic posterior-sampling algorithm for the SSP setting. The algorithm operates in epochs and uses a logarithmic, time-dependent schedule for the number of posterior samples. To induce optimism, we augment the transition model with a pseudo-state assigned an optimistic boundary value. We establish a high-probability regret bound of $\tilde{\mathcal{O}}(B_\star \sqrt{SAK} + B_\star S^2 A)$, where $B_\star$ is an upper bound on the expected cost of the optimal policy, $S$ and $A$ are the numbers of states and actions, respectively, and $K$ is the number of episodes. 
In addition, we extend the analysis to general zero-cost SSPs using a universal cost-perturbation argument. The leading term matches the minimax lower bound $\Omega(B_\star \sqrt{SAK})$, thereby addressing the open problem raised by \citet{jafarniajahromi2021onlinelearningstochasticshortest} for posterior sampling in the SSP setting.
\end{abstract}

\addtocontents{toc}{\protect\setcounter{tocdepth}{-1}}

\section{Introduction}

The stochastic shortest path (SSP) problem \citep{tarbouriech2020noregretexplorationgoalorientedreinforcement} is a framework for goal-oriented reinforcement learning (RL), in which an agent seeks to reach a designated terminal state while minimizing the expected cumulative cost. SSP generalizes classical RL settings, including finite-horizon episodic Markov decision processes (MDPs) and discounted MDPs. Specifically, finite-horizon MDPs can be formulated as SSPs by augmenting the state space with time indices and enforcing termination after $H$ steps. Discounted MDPs can be embedded into SSPs through a geometric termination mechanism, in which the discount factor corresponds to the probability of continuation. Thus, SSP provides a common framework for fixed-horizon, randomly terminated, and goal-reaching RL problems.

This generality, however, makes learning in unknown SSPs particularly challenging. Since the hitting time is policy-dependent and may be unbounded under improper policies, exploration must be carefully designed to avoid non-terminating behavior. Moreover, the high variance in value estimates induced by random hitting times further complicates the design of exploration strategies with provable regret guarantees \citep{tarbouriech2020noregretexplorationgoalorientedreinforcement}.

In recent years, a common approach to this challenge has been Optimism in the Face of Uncertainty (OFU) \citep{lai1985asymptotically}.
A series of studies \citep{tarbouriech2020noregretexplorationgoalorientedreinforcement, cohen2020nearoptimalregretboundsstochastic, cohen2021minimaxregretstochasticshortest, tarbouriech2021stochasticshortestpathminimax, chen2021implicitfinitehorizonapproximationefficient,rosenberg2022stochasticshortestpathadversarially,chen2021findingstochasticshortestpath} has progressively refined OFU-based algorithms, culminating in the minimax-optimal leading regret term of $\tilde{\mathcal{O}}(B_\star \sqrt{SAK})$, where $B_\star$ is an upper bound on the expected cost of the optimal policy, $S$ and $A$ are the numbers of states and actions, respectively, and $K$ is the number of episodes. Table~\ref{tab:regret_comparison} provides a detailed comparison. Here, ``horizon-free'' follows the standard SSP convention of excluding polynomial dependence on the optimal-policy expected hitting time.

Despite their strong theoretical guarantees, OFU-based algorithms typically require explicit variance-aware optimism, implemented through confidence sets or exploration bonuses. Constructing and maintaining these uncertainty estimates, together with planning under the resulting optimistic models, can impose substantial computational overhead in model-based SSP algorithms.
Posterior sampling (PS), also known as Thompson sampling~\citep{thompson1933likelihood}, offers a simpler and empirically competitive alternative.
Recently, \citet{jafarniajahromi2021onlinelearningstochasticshortest} extended PS to the SSP setting by proposing PSRL-SSP, demonstrating strong empirical performance against existing OFU-type algorithms.
However, its regret bound incurs an additional $\sqrt{S}$ factor relative to the minimax lower bound. This gap leads to an important open question, highlighted by \citet{jafarniajahromi2021onlinelearningstochasticshortest}:

\begin{center}
    \textbf{Can posterior-sampling-based algorithms achieve minimax-optimal regret in SSP?}
\end{center}

Inspired by Optimistic Posterior Sampling (OPSRL) for finite-horizon episodic MDPs \citep{tiapkin2022optimisticposteriorsamplingreinforcement}, we answer this question in the affirmative by proposing OPSRL-SSP, the first optimistic posterior-sampling algorithm for the SSP setting. However, extending OPSRL from finite-horizon reward maximization to SSPs presents two challenges. First, the absence of a fixed horizon $H$ rules out static sampling schedules that rely on a known time horizon. Second, moving from reward maximization to cost minimization requires a different construction of optimistic priors.

To address these challenges, OPSRL-SSP incorporates optimism into posterior sampling through three mechanisms. First, we augment the state space with a pseudo-state $s_0$ assigned an optimistically low boundary value. This provides a virtual boundary condition for stationary cost-minimizing planning. Second, to maintain optimism at data-dependent epoch starts without a finite time horizon, we use a logarithmic, time-dependent number of posterior candidates $J_{t_\ell}$. This schedule makes the corresponding failure probabilities summable over global time. Third, we introduce an adaptive inflation parameter $\kappa_{t_\ell}(s,a)$ for the Dirichlet pseudo-counts, which maintains a constant lower bound on the probability of an optimistic transition candidate for each state-action pair. Together, these mechanisms ensure with high probability that each state-action pair has an optimistic transition candidate at every epoch, enabling the planner to compute an optimistic value function. Our main contributions are as follows:

\begin{itemize}[leftmargin=*]
    \item We propose OPSRL-SSP, the first optimistic posterior sampling algorithm for stochastic shortest path problems. By combining a pseudo-state construction, posterior inflation, and an anytime sampling schedule, it extends optimistic posterior sampling to the challenging unbounded-horizon SSP setting.

    \item We establish a global anytime optimism guarantee and prove a regret bound with minimax-optimal leading term $\tilde{\mathcal{O}}(B_\star \sqrt{SAK})$. The analysis includes a variance-sensitive perturbation argument with a Freedman-based self-bounding technique. We further extend our method to zero-cost SSPs via cost perturbation.

    \item We empirically evaluate OPSRL-SSP on standard tabular SSP benchmarks, including RandomMDP and GridWorld, as well as a held-out RandomMDP cohort. Across these experiments, OPSRL-SSP remains competitive with PSRL-SSP and top-performing OFU baselines, while substantially outperforming several less efficient baselines.
\end{itemize}

\begin{table}[!t]
\centering
\caption{Comparison of state-of-the-art regret bounds for tabular SSP. Here, $S$ denotes the state-space size, $A$ denotes the action-space size, and $K$ the number of episodes. $B_\star$ denotes an upper bound on $\max_{s\in\mathcal S}V^\star(s)$. $D$ is the SSP diameter, $c_{\min}$ is the minimum instantaneous cost, and $T_\star$ denotes the maximum expected hitting time of the optimal policy. All bounds suppress polylogarithmic factors. ``Horizon-free'' means that the regret bound depends on $T_\star$ at most logarithmically. In positive-cost settings, this also rules out polynomial dependence on $B_\star/c_{\min}$. For OPSRL-SSP, this designation refers to the strictly positive-cost result in Theorem~1.}
\label{tab:regret_comparison}
\resizebox{\textwidth}{!}{%
\begin{tabular}{@{}lclc@{}}
\toprule
\textbf{Algorithm} & \textbf{Approach} & \textbf{Regret Bound} & \textbf{Horizon-Free} \\
\midrule
UC-SSP \citep{tarbouriech2020noregretexplorationgoalorientedreinforcement}  & OFU & $\tilde{\mathcal{O}}(D S \sqrt{D A K / c_{\min}} + S^2 A D^2)$ & No \\
Bernstein-SSP \citep{cohen2020nearoptimalregretboundsstochastic}            & OFU & $\tilde{\mathcal{O}}(B_\star S \sqrt{A K} + \sqrt{B_\star^3 S^2 A^2 / c_{\min}})$ & No \\
ULCVI \citep{cohen2021minimaxregretstochasticshortest}                      & OFU & $\tilde{\mathcal{O}}(B_\star \sqrt{S A K} + T_\star^4 S^2 A)$ & No \\
EB-SSP \citep{tarbouriech2021stochasticshortestpathminimax}                 & OFU & $\tilde{\mathcal{O}}(B_\star \sqrt{S A K} + B_\star S^2 A)$ & \textbf{Yes} \\
LCB-ADVANTAGE-SSP \citep{chen2021implicitfinitehorizonapproximationefficient} & OFU & $\tilde{\mathcal{O}}(B_\star \sqrt{S A K} + B_\star^5 S^2 A / c_{\min}^4)$ & No \\
SVI-SSP \citep{chen2021implicitfinitehorizonapproximationefficient}         & OFU & $\tilde{\mathcal{O}}(B_\star \sqrt{S A K} + B_\star S^2 A)$ & \textbf{Yes} \\
PSRL-SSP \citep{jafarniajahromi2021onlinelearningstochasticshortest}        & PS  & $\tilde{\mathcal{O}}(B_\star S \sqrt{A K} + S^2 A \sqrt{B_\star^3 / c_{\min}})$ & No \\
Lower Bound \citep{cohen2020nearoptimalregretboundsstochastic} & -- & $\Omega(B_\star \sqrt{S A K})$ & -- \\
\midrule
\textbf{OPSRL-SSP (Ours; Theorem~1)} & \textbf{OPS} & $\tilde{\mathcal{O}}(B_\star \sqrt{S A K} + B_\star S^2 A)$ & \textbf{Yes} \\
\bottomrule
\end{tabular}%
}
\end{table}

\paragraph{Related Work}
Stochastic shortest path problems have received substantial attention. Table~1 of \citet{chen2021implicitfinitehorizonapproximationefficient} and Table~\ref{tab:regret_comparison} compare representative results for tabular SSPs. UC-SSP \citep{tarbouriech2020noregretexplorationgoalorientedreinforcement} established early regret guarantees, but its bound depends on the SSP diameter. Subsequent OFU-based methods improved this dependence. Bernstein-SSP \citep{cohen2020nearoptimalregretboundsstochastic} removed the diameter dependence, while ULCVI \citep{cohen2021minimaxregretstochasticshortest} achieved the optimal leading term with a relatively large lower-order term. EB-SSP \citep{tarbouriech2021stochasticshortestpathminimax} and SVI-SSP \citep{chen2021implicitfinitehorizonapproximationefficient} obtained minimax-optimal, horizon-free guarantees with matching lower-order complexity. By contrast, the regret bound for PSRL-SSP \citep{jafarniajahromi2021onlinelearningstochasticshortest} retains an additional $\sqrt{S}$ factor relative to the minimax lower bound. Beyond the standard tabular setting, studies have considered SSPs with adversarially changing costs \citep{rosenberg2022stochasticshortestpathadversarially,chen2021findingstochasticshortestpath}, linear function approximation \citep{vial2021regretboundsstochasticshortest,min2022learningstochasticshortestpath}, generative models \citep{chen2022reachinggoalshardsettling,pmlr-v132-tarbouriech21a}, and decentralized multi-agent interaction \citep{chavan2025regretlowerboundsdecentralized}. Within the function-approximation literature, \citet{di2024nearlyminimaxoptimalregret} obtain a nearly minimax-optimal regret guarantee for linear-mixture SSPs, while \citet{polikar2025regretguaranteeslinearcontextual} introduce linear contextual SSPs with adversarially chosen contexts.

Posterior-sampling methods have been studied in bandits and reinforcement learning. Early RL methods had suboptimal regret guarantees \citep{osband2013moreefficientreinforcementlearning}, while subsequent work introduced optimism through several mechanisms. In contextual bandits, Feel-Good Thompson Sampling tilts the posterior toward models with larger predicted rewards; later variants incorporate variance-dependent weighting \citep{zhang2021feelgoodthompsonsamplingcontextual,li2025varianceawarefeelgoodthompsonsampling}. \citet{agarwal2022modelbasedrloptimisticposterior} extend this value-tilting principle to model-based RL by incorporating predicted optimal values into the likelihood. In average-reward MDPs, \citet{agrawal2020posterior} introduce an optimistic posterior-sampling component based on selecting among multiple draws from an inflated Dirichlet posterior. For finite-horizon episodic MDPs, optimistic Thompson-sampling methods reshape posterior samples by clipping pessimistic values and increasing posterior variance. OPSRL instead draws transition candidates from an inflated Dirichlet posterior and selects optimistically from a logarithmic number of candidates \citep{hu2023optimistic,tiapkin2022optimisticposteriorsamplingreinforcement}. Other work develops model-free posterior sampling with optimistic priors under general structural conditions \citep{dann2022provablyefficientmodelfreeposterior}. Related extensions study posterior sampling in Gaussian-process MDPs \citep{bayrooti2025noregretthompsonsamplingfinitehorizon}, prior-aligned meta-reinforcement learning \citep{zhou2025prioralignedmetarlthompsonsampling}, and concurrent multi-agent learning \citep{NEURIPS2022_32133a6a}. Parallel work investigates horizon-free guarantees \citep{zhou2022computationallyefficienthorizonfreereinforcement} and alternative computational approaches, such as quantum value iteration \citep{luo2025quantumalgorithmsfinitehorizonmarkov}.

\section{Preliminaries}

\subsection{Stochastic Shortest Path}
We consider an unknown stochastic shortest path (SSP) problem defined by the tuple $\mathcal{M} = (\mathcal{S}, \mathcal{A}, p, c, s_{init}, g)$. The finite state space $\mathcal{S}$ has size $S$, the finite action space $\mathcal{A}$ has size $A$, $s_{init} \in \mathcal{S}$ is a fixed initial state, and $g \notin \mathcal{S}$ is the unique absorbing goal state. The transition kernel $p: \mathcal{S}^+ \times \mathcal{S} \times \mathcal{A} \to [0, 1]$ governs the physical dynamics: $p(s'|s, a) = \mathbb{P}(s_{t+1} = s' \mid s_t = s, a_t = a)$, where $\mathcal{S}^+ = \mathcal{S} \cup \{g\}$ is the physical state space including the goal. At time step $t = 1,2,\ldots$, $s_t \in \mathcal{S}$ and $a_t \in \mathcal{A}$ denote the state and action, and $s_{t+1} \in \mathcal{S}^+$ is the next state. The known cost function $c: \mathcal{S} \times \mathcal{A} \to [0, 1]$ assigns a cost to each state-action pair. The episode terminates upon reaching $g$, where $p(g|g,a)=1$ and $c(g,a)=0$ for all $a\in\mathcal{A}$. We define the minimum instantaneous cost as $c_{\min}=\min_{s,a}c(s,a)$ and allow zero-cost transitions (that is, $c_{\min}\ge 0$); see Appendix~\ref{sec:appendix_c_cost_perturbation}.

\paragraph{Policies and Bellman Equations}
A deterministic stationary policy $\pi: \mathcal{S} \to \mathcal{A}$ maps each state to an action. The value function (also referred to as the cost-to-go function) associated with a policy $\pi$ is a mapping $V^\pi: \mathcal{S}^+ \to [0, \infty)$ given by $V^\pi(g) = 0$ and
\[
V^\pi(s) = \mathbb{E}\left[\sum_{t=1}^{T} c(s_t, \pi(s_t)) \mid s_1 = s\right], \quad \forall s \in \mathcal{S}
\]
where $T = \min\{t \ge 1: s_t = g\}$ is the random stopping time denoting the number of steps before reaching the goal state when policy $\pi$ is followed. A policy $\pi$ is called \textit{proper} if, starting from any initial state, the goal state is reached with probability 1 (i.e., $T < \infty$ almost surely). Otherwise, it is called improper. For any \textit{proper} policy $\pi$, its state-action value function (or $Q$-function) is defined as:
\[
Q^\pi(s, a) = c(s, a) + \sum_{s' \in \mathcal{S}^+} p(s'|s, a) V^\pi(s')
\]
and its state value function satisfies the Bellman equation $V^\pi(s) = Q^\pi(s, \pi(s))$.
The optimal value function is defined as $V^\star(s) = \min_\pi V^\pi(s)$. Accordingly, the optimal $Q$-function and $V$-function are the unique solutions to the Bellman optimality equations \citep{bertsekas1991analysis}:
\[
Q^\star(s, a) = c(s, a) + \sum_{s' \in \mathcal{S}^+} p(s'|s, a) V^\star(s'), \quad \text{and} \quad V^\star(s) = \min_{a \in \mathcal{A}} Q^\star(s, a).
\]
When the transition kernel $p$ is known, computing the optimal policy is a planning problem that can be solved using value iteration or policy iteration \citep{bertsekas2017dynamic}. In our online learning setting, the agent knows the state space $\mathcal{S}$, action space $\mathcal{A}$, and cost function $c$, but not the transition kernel $p$. We address this uncertainty using a Bayesian framework in which the transition model is drawn from a prior distribution $\mu_0$.

\subsection{Assumptions and Regret Definition}
Following standard SSP literature \citep{cohen2021minimaxregretstochasticshortest, tarbouriech2021stochasticshortestpathminimax}, we make the following assumption.
\begin{assumption}\label{ass:ssp}
For each SSP $\mathcal M$ with transition kernel $p$, the following holds: (1) there exists at least one proper policy; (2) for every improper policy $\pi$, there exists a state $s\in\mathcal S$ such that $V^\pi(s)=+\infty$; and (3) $B_\star\ge1$ and the optimal value function satisfies $\max_{s\in\mathcal S}V^\star(s)\le B_\star$.
\end{assumption}

\paragraph{Learning Protocol and Regret}
The agent interacts with the unknown environment for $K$ episodes, each of which begins at $s_{init}$. Since SSP episode lengths are random and potentially unbounded, our algorithm updates its policy at epoch boundaries determined by the count-doubling and episode-increment criteria. We measure performance by the cumulative regret $R_K$ over $K$ episodes, defined as the difference between the realized cumulative cost and the expected cost of the optimal policy:
\[
R_K = \sum_{k=1}^K \sum_{t=1}^{I_k} c(s_{k, t}, a_{k, t}) - K \cdot V^\star(s_{init})
\]
where $I_k$ denotes the number of steps in episode $k$. The objective is to minimize cumulative regret.

\subsection{Empirical Counts and Posterior Model}

At global time step $t$, let $n_t(s,a)=\sum_{i=1}^{t-1}\mathbbm{1}\{(s_i,a_i)=(s,a)\}$ be the number of visits to state-action pair $(s,a)$. Let $n_t(s'|s,a)=\sum_{i=1}^{t-1}\mathbbm{1}\{(s_i,a_i,s_{i+1})=(s,a,s')\}$ denote the number of observed transitions from $s$ to $s'$ after action $a$. We define the empirical transition distribution by $\widehat{p}_t(s'|s,a)=n_t(s'|s,a)/n_t(s,a)$.

\paragraph{Posterior Sampling and Improper Dirichlet Prior}
To address the exploration-exploitation trade-off, we use the posterior sampling (PS) framework. In standard PS, the agent updates a prior distribution over the unknown transition kernel $p$ using empirical transition counts $N(s'|s,a)$. Since the transition vector $\bigl(p(s'|s,a)\bigr)_{s'\in\mathcal S^+}$ lies on the probability simplex $\Delta(\mathcal{S}^+)$, the Dirichlet distribution is a natural conjugate prior.

To achieve minimax optimality, we follow \citet{tiapkin2022optimisticposteriorsamplingreinforcement} and allow improper Dirichlet parameters, i.e., zero pseudo-counts on some coordinates. Concretely, for any nonnegative parameter vector $\alpha=(\alpha_{s'})_{s'\in\mathcal{X}} \in \mathbb{R}_+^{|\mathcal{X}|}$ with $\sum_{s'\in\mathcal{X}}\alpha_{s'}>0$, we define $\mathrm{Dir}(\alpha)$ on $\Delta(\mathcal{X})$ as follows: let
\[
\mathcal{I}(\alpha)\triangleq \{s'\in\mathcal{X}:\alpha_{s'}>0\}.
\]
We define a random vector $P\sim\mathrm{Dir}(\alpha)$ by first sampling
$
(P_{s'})_{s'\in\mathcal{I}(\alpha)} \sim \mathrm{Dir}\big((\alpha_{s'})_{s'\in\mathcal{I}(\alpha)}\big)
$
on the lower-dimensional simplex over $\mathcal{I}(\alpha)$, and then setting $P_{s'}=0$ for all $s'\notin\mathcal{I}(\alpha)$. In this sense, zero-parameter coordinates are deterministically fixed to zero, while the strictly positive coordinates follow an ordinary proper Dirichlet law.

After observing transition counts $N(s'|s,a)$, the posterior distribution is given by
\[
\mathrm{Dir}\big((\alpha_0(s'|s,a)+N(s'|s,a))_{s'\in\mathcal{S}^+}\big)
\]
using the extended definition above. This formulation accommodates zero pseudo-counts and allows optimistic pseudo-counts to be assigned only to the designated pseudo-state, providing the optimism mechanism used below.

\section{The OPSRL-SSP Algorithm}

\subsection{Algorithm Design}
Algorithm~\ref{alg:opsrl_ssp} presents OPSRL-SSP, which combines posterior sampling with the following SSP-specific mechanisms.

\begin{algorithm}[!t]
\small
\DontPrintSemicolon
\SetInd{0.5em}{0.7em}
\SetNlSty{textnormal}{}{}
\SetAlgoNlRelativeSize{-1}
\caption{\texttt{OPSRL-SSP}}
\label{alg:opsrl_ssp}

\textbf{Initialization:} $t \gets 1, \ell \gets 1, K_0 \gets 0, t_1 \gets 1, k_1 \gets 1$\;
Initialize visit and transition counts as zero, and initialize the first epoch by injecting $n_{0,t_1}(s,a)$, sampling $\{\bigl(\tilde p_1^j(s'|s,a)\bigr)_{s'\in\tilde{\mathcal S}}\}_{j\in[J_{t_1}]}$ for all $(s,a)$, and solving for $\tilde V_1,\pi_1$\;
\For{episodes $k = 1, 2, \ldots, K$}{
    $s_t \gets s_{\text{init}}$\;
    \While{$s_t \neq g$}{
        \If{$k - k_\ell > K_{\ell-1}$ \textbf{or} $n_t(s, a) > 2n_{t_\ell}(s, a)$ \textit{for some} $(s, a) \in \mathcal{S} \times \mathcal{A}$}{
            $K_\ell \gets k - k_\ell$, $\ell \gets \ell + 1$,  $t_\ell \gets t$,  $k_\ell \gets k$\;
            \For{all $(s, a) \in \mathcal{S} \times \mathcal{A}$}{
                $n_{t_\ell}(s, a) \gets n_t(s, a)$\;
                $\bar{n}_{t_\ell}(s'|s, a) \gets n_t(s'|s, a) + \mathbbm{1}\{s' = s_0\} \cdot n_{0,t_\ell}(s,a), \quad \forall s' \in \tilde{\mathcal{S}}$\;
            }
            For all $(s, a) \in \mathcal{S} \times \mathcal{A}$, sample $J_{t_\ell}$ independent transition vectors over $\tilde{\mathcal{S}} = \mathcal{S}^+ \cup \{s_0\}$:\;
            \[ \bigl(\tilde{p}_{\ell}^{j}(s'|s, a)\bigr)_{s'\in\tilde{\mathcal S}} \sim \operatorname{Dir}\!\left(\left(
            \frac{\bar{n}_{t_\ell}(s'|s, a)}{\kappa_{t_\ell}(s, a)}\right)_{s' \in \tilde{\mathcal{S}}}\right),
            \quad j \in [J_{t_\ell}]\]
            Set $\overline V_\ell^{(m)}(g)\gets0$ and $\overline V_\ell^{(m)}(s_0)\gets-2B_\star$ at each iteration; repeat until convergence:\;
            $\displaystyle \begin{aligned}
                Q_\ell^{(m+1)}(s,a)&\gets c(s,a)+\min_{j\in[J_{t_\ell}]}\sum_{x\in\tilde{\mathcal S}}\tilde p_\ell^j(x|s,a)\overline V_\ell^{(m)}(x),\\
                V_\ell^{(m+1)}(s)&\gets\min_{a\in\mathcal A}Q_\ell^{(m+1)}(s,a).
            \end{aligned}$\;
            Set $\tilde V_\ell\gets\lim_m V_\ell^{(m)}$, $\tilde Q_\ell\gets\lim_m Q_\ell^{(m)}$, and $\pi_\ell(s)\in\argmin_a\tilde Q_\ell(s,a)$\;
        }
        Choose action $a_t = \pi_\ell(s_t)$ and execute it\;
        Observe actual next state $s_{t+1} \sim \bigl(p(s'|s_t,a_t)\bigr)_{s'\in\mathcal S^+}$\;
        $n_{t+1}(s_t, a_t) \gets n_t(s_t, a_t) + 1$\;
        $n_{t+1}(s_{t+1}|s_t, a_t) \gets n_t(s_{t+1}|s_t, a_t) + 1$\;
        $t \gets t + 1$\;
    }
}
\end{algorithm}

We use the following parameter schedules to establish the theoretical guarantee. For a global time $t$ and visit count $n$, let
\[
\beta^\star(\delta,n)=\log(12SA/\delta)+3\log\!\bigl(e\pi(2n+1)\bigr),
\qquad \kappa_t(s,a)=2\beta^\star\!\bigl(\delta,n_t(s,a)\bigr).
\]
The pseudo-count and the number of independent candidates are
\[
n_{0,t}(s,a)=\left\lceil\kappa_t(s,a)\left[c_0+\log_{17/16}\!\left(\frac{n_t(s,a)}{\kappa_t(s,a)}+1\right)\right]\right\rceil,
\qquad
J_t=\left\lceil\frac{\log(4SA\pi^2t^2/\delta)}{-\log(1-\gamma)}\right\rceil,
\]
where $c_0$ and the absolute one-sample optimism probability $\gamma>0$ are specified in Proposition~\ref{prop:optimism_ssp}.


\paragraph{Epoch-based Execution and Dual-Stopping Criterion}
As in PSRL-SSP \citep{jafarniajahromi2021onlinelearningstochasticshortest}, OPSRL-SSP operates in epochs and uses two stopping conditions to accommodate random and potentially unbounded episode lengths. At the beginning of each epoch $\ell$, the algorithm samples candidate transition rows for every state-action pair, computes a stationary policy from these candidates, and follows that policy until either stopping condition is met.
The first is a count-doubling event: for some $(s,a)\in\mathcal S\times\mathcal A$,
\[
n_t(s,a)>2n_{t_\ell}(s,a),
\]
where $t_\ell$ is the starting time of epoch $\ell$. The second condition is triggered by an increase in the number of episodes:
\[
k-k_\ell>K_{\ell-1},
\]
where $k_\ell$ is the starting episode of epoch $\ell$ and $K_{\ell-1}$ is the number of episodes in the previous epoch. Together, the two conditions limit epoch duration and control the number of epochs in the regret analysis.

\paragraph{Optimistic Pseudo-state} In the cost-minimizing SSP setting, we augment the physical state space with a virtual pseudo-state $s_0$, forming $\tilde{\mathcal{S}}=\mathcal{S}^+\cup\{s_0\}$. Earlier optimistic pseudo-state constructions \citep{Brafman02RMAX,szita2008} motivate this augmentation. We assign $s_0$ the internal boundary value $-2B_\star$. This value is used for planning and does not correspond to a state that a policy must reach in the original SSP. It serves two purposes. For $f=B_\star-\bar V^\star$, it gives $f(s_0)=3B_\star$, the separated coordinate used in the Dirichlet anti-concentration argument. For the shifted value $U_\ell=\tilde V_\ell+2B_\star$, it gives $U_\ell(s_0)=0$, which simplifies the analysis of pseudo-count bias. These arguments are given in Appendices~\ref{sec:app_b2_optimism} and~\ref{sec:app_b3_regret}, respectively.

\paragraph{Pseudo-counts and Extended Transitions}
At the start of each epoch $\ell$ (global step $t_\ell$), we introduce an optimistic pseudo-count $n_{0,t_\ell}(s, a)$ for each state-action pair. The shifted total count is defined as $\bar{n}_{t_\ell}(s, a) = n_{t_\ell}(s, a) + n_{0,t_\ell}(s, a)$. These injected pseudo-counts are exclusively allocated to the virtual transition pointing to $s_0$. The extended transition counts over the simulated space $\tilde{\mathcal{S}}$ are thus formulated as:
\[
\bar{n}_{t_\ell}(s'|s, a) = n_{t_\ell}(s'|s, a) + n_{0,t_\ell}(s, a) \cdot \mathbbm{1}\{s' = s_0\}, \quad \forall s' \in \tilde{\mathcal{S}}.
\]
Accordingly, we define the pseudo-empirical transition probability distribution as $\bar{p}_{t_\ell}(s'|s, a) = \bar{n}_{t_\ell}(s'|s, a) / \bar{n}_{t_\ell}(s, a)$.

\paragraph{Dynamic Posterior Inflation} At the start of epoch $\ell$, we divide the augmented Dirichlet counts by the state-action- and visit-count-dependent factor $\kappa_{t_\ell}(s,a)$ before sampling. This preserves the mean of the augmented Dirichlet row while reducing its concentration and increasing the probability of an optimistic lower-tail draw. The factor $\kappa_t$ is determined by the time-uniform lower-tail KL confidence budget $\beta^\star$. Specifically, for each $(s,a)$ and $j\in[J_{t_\ell}]$, we draw
\[
\bigl(\tilde{p}_\ell^j(s'|s, a)\bigr)_{s'\in\tilde{\mathcal S}} \sim \operatorname{Dir}\left(\left(\frac{\bar{n}_{t_\ell}(s'|s, a)}{\kappa_{t_\ell}(s, a)}\right)_{s' \in \tilde{\mathcal{S}}}\right).
\]
The analysis below shows that this choice gives a uniform positive lower bound on the one-sample optimism probability.

\paragraph{Anytime Optimistic Sampling and Internal Planning}
OPSRL-SSP draws $J_{t_\ell}=\Theta(\log(SAt_\ell/\delta))$ independent candidates at each epoch. This choice makes the row-wise optimism failure probabilities summable over all global times, so the event also holds at data-dependent epoch starts. The agent then solves the stationary Bellman fixed-point problem on $\tilde{\mathcal S}$, taking the minimum predicted cost over the candidates for each state-action pair. One Bellman sweep requires $O(J_{t_\ell}S^2A)$ transition-coordinate operations.

Together, the pseudo-state, dynamic posterior inflation, and anytime sampling turn row-wise optimistic candidates into an optimistic stationary value function; the formal argument is outlined in Section~\ref{sec:proof_sketch_theorem_one}.

\subsection{Regret Analysis}
We first present the regret guarantee of our OPSRL-SSP algorithm under the strictly positive-cost setting.

\textbf{Theorem 1 (Regret Bound of OPSRL-SSP)} \emph{Suppose Assumption~\ref{ass:ssp} holds, an upper bound $B_\star \ge 1$ on the optimal expected cost is known, and the minimum instantaneous cost satisfies $c_{\min} > 0$. Consider Algorithm~\ref{alg:opsrl_ssp} instantiated with appropriately chosen logarithmic posterior sample counts \(J_t\), state-action-dependent inflation parameters \(\kappa_t(s,a)\), and pseudo-counts \(n_{0,t}(s,a)\) as specified in Proposition~\ref{prop:optimism_ssp}. Then, for any $\delta \in (0,1)$, with probability at least $1-\delta$, the cumulative regret $R_K$ over $K$ episodes satisfies}
\[
R_K \le \tilde{\mathcal{O}}\left( B_\star \sqrt{SAK} + B_\star S^2 A \right).
\]

\textbf{Remark 1 (Minimax Optimality)}
Theorem 1 matches the minimax lower bound $\Omega(B_\star\sqrt{SAK})$ in the leading term.
In particular, OPSRL-SSP removes the extra $\sqrt{S}$ factor in the regret bound of PSRL-SSP by enforcing anytime optimism.

\textbf{Remark 2 (Lower-Order Complexity)}
The lower-order state-complexity term of our bound, $B_\star S^2 A$, matches the state-of-the-art lower-order complexity achieved by OFU-style algorithms such as EB-SSP \citep{tarbouriech2021stochasticshortestpathminimax} and SVI-SSP \citep{chen2021implicitfinitehorizonapproximationefficient}. This work shows that, by employing a variance-sensitive perturbation analysis, posterior sampling incurs no extra state-action penalty and yields the same lower-order complexity as the best existing tabular OFU methods.

\paragraph{Extension to Zero-Cost SSPs} To extend our analysis beyond the strictly positive-cost setting, we study OPSRL-SSP under a universal cost perturbation \citep{tarbouriech2021stochasticshortestpathminimax,chen2021implicitfinitehorizonapproximationefficient}; the corresponding algorithm is presented in Appendix~\ref{sec:appendix_c_cost_perturbation}. The perturbation raises every cost to at least $\epsilon$, so the perturbed problem satisfies the strictly positive-cost condition required by the base analysis. This gives the following regret guarantee for zero-cost SSPs.

\textbf{Theorem 2 (Cost-Perturbed Regret Bound for Zero-Cost SSPs)}
\emph{Suppose Assumption~\ref{ass:ssp} holds, and assume the costs are known and nonnegative. Fix a perturbation level $0<\epsilon\le1$, and define the perturbed cost function $c_\epsilon(s,a)=\max\{c(s,a),\epsilon\}$. Let $\mathcal M_\epsilon$ denote the corresponding perturbed SSP, and suppose Algorithm~\ref{alg:opsrl_ssp} is executed on $\mathcal M_\epsilon$ with any valid upper bound $B_\epsilon \ge \max\{1,B_\star^\epsilon\}$, where $B_\star^\epsilon \triangleq \max_{s\in\mathcal S} V_\epsilon^\star(s)$. Let $T_\star$ denote the maximum expected hitting time of an optimal policy in the original SSP. Then, with probability at least $1-\delta$, its cumulative regret in the original SSP satisfies}
\[
R_K\le\tilde{\mathcal O}\!\left(B_\epsilon \sqrt{SAK}+B_\epsilon S^2A\right)+K\epsilon T_\star.
\]

\subsection{Proof Sketch for Theorem 1}
\label{sec:proof_sketch_theorem_one}

This section sketches the analysis of Theorem 1. Complete proofs are given in Appendix~\ref{sec:appendix_b_proofs}. The analysis first establishes global optimism and then bounds regret in three steps.

\paragraph{The Foundation of Global Anytime Optimism.}
The first step of the analysis is to show that the value function computed from the sampled models is optimistic uniformly over all epochs, namely $\tilde V_\ell(s)\le V^\star(s)$ for all $s\in\mathcal S$.
This is a central challenge in extending optimistic posterior sampling to SSPs, where episode lengths are random and may be unbounded.

Our proof proceeds in two stages.
First, we translate optimism into a one-step lower-tail event,
\[
\min_{j\in[J_{t_\ell}]}\tilde p_\ell^j \bar V^\star(s,a)\le pV^\star(s,a),
\]
where $\bar V^\star$ extends $V^\star$ to the pseudo-state by setting $\bar V^\star(s_0)=-2B_\star$.
The pseudo-state converts lower-cost optimism into an upper-tail event for $f=B_\star-\bar V^\star$. Its value $f(s_0)=3B_\star$ supplies the separated coordinate used in the Gaussian lower-bound lemma for Dirichlet distributions (Lemma~\ref{lem:dirichlet_anti_concentration}). On the time-uniform lower-tail KL event established in Appendix~\ref{sec:app_b1_concentration}, the comparison in the proof of Proposition~\ref{prop:optimism_ssp} bounds the augmented row's count-weighted lower-tail KL term by $\beta^\star(\delta,n_{t_\ell}(s,a))$. With $\kappa_{t_\ell}(s,a)=2\beta^\star(\delta,n_{t_\ell}(s,a))$, the Gaussian-tail argument is bounded by a universal constant. The anti-concentration argument in that proof then gives a uniform positive probability that a candidate is optimistic. Drawing $J_{t_\ell}$ independent candidates reduces the row-wise failure probability to order $t_\ell^{-2}$. A union bound over state-action pairs and deterministic global times yields the anytime anti-concentration event, including at data-dependent epoch starts.

Second, we lift this one-step optimism to global value-function optimism.
The injected pseudo-count gives every sampled model a positive probability of transitioning to $s_0$, making the optimistic Bellman operator over the augmented space a strict contraction.
Under the one-step anti-concentration event, $V^\star$ is a supersolution of this operator. Monotonicity and contraction then imply that its unique fixed point satisfies $\tilde V_\ell\le V^\star$ for every epoch.
This establishes the global anytime optimism event
\[
\mathcal E^{\mathrm{opt}}\triangleq\left\{\tilde V_\ell(s)\le V^\star(s)\text{ for every epoch }\ell\text{ and every }s\in\mathcal S\right\},
\]
used in the regret decomposition (see Appendix~\ref{sec:app_b2_optimism} for the formal proof).

We now bound the cumulative regret $R_K$ on the event $\mathcal{E}^{\mathrm{opt}}$ in three steps:

\paragraph{Step 1: Regret Decomposition and MDS Construction.}
Let $T_K$ denote the total number of time steps in the first $K$ episodes. We rewrite $R_K$ as a sum over global time steps $\tau$. On $\mathcal{E}^{\mathrm{opt}}$, optimism lets us replace the optimal baseline value with $\tilde{V}_{\ell}(s_{init})$. We then use the optimistic Bellman equation and the true transition expectation to decompose the regret. To handle potentially negative values of $\tilde{V}_\ell$ induced by the pseudo-state, we introduce the nonnegative shifted value function $U_\ell = \tilde{V}_\ell + 2B_\star \in [0, 3B_\star]$ (see Appendix~\ref{sec:app_b3_regret} for the derivation):
\[
R_K \le \sum_{\tau=1}^{T_K} \widetilde{M}_\tau + \sum_{\tau=1}^{T_K} \mathrm{Error}_\tau + R_{\mathrm{trunc}}.
\]
Here, $\widetilde{M}_\tau := U_{\ell(\tau)}(s_{\tau+1}) - p(\cdot|s_\tau, a_\tau)U_{\ell(\tau)}$ forms a Martingale Difference Sequence (MDS) with zero conditional expectation. The term $\mathrm{Error}_\tau := p(\cdot|s_\tau, a_\tau)U_{\ell(\tau)} - \min_{j\in[J_{t_\ell}]} \tilde{p}^j_{\ell(\tau)}U_{\ell(\tau)}$ captures the single-step model estimation error.
Finally, \(R_{\mathrm{trunc}}\) accounts for the telescoping residual across epoch switches. On $\mathcal E^{\mathrm{opt}}$, $-2B_\star\le\tilde V_\ell(s)\le V^\star(s)\le B_\star$, so the jump between two consecutive internal value functions at a switch is at most $3B_\star$. By Lemma~\ref{lem:number_epochs}, the total number of epochs \(L_M\) up to the completion of \(K\) episodes satisfies \(L_M=\tilde{\mathcal O}(\sqrt{SAK}+SA)\). Thus \(R_{\mathrm{trunc}}\le3B_\star L_M=\tilde{\mathcal O}(B_\star\sqrt{SAK}+B_\star SA)\).

\paragraph{Step 2: Statistical Decomposition of Model Error.}
Let $\bar{p}_{t_\ell}$ denote the extended empirical distribution that includes the pseudo-counts. We decompose $\mathrm{Error}_\tau$ into the empirical concentration error $\xi^{\mathrm{conc}}_\tau$, the pseudo-count bias $\xi^{\mathrm{bias}}_\tau$, and the optimistic posterior deviation $\xi^{\mathrm{opt}}_\tau$, all evaluated at the shifted value function $U_\ell$:

\[
\mathrm{Error}_\tau = \underbrace{(p - \hat{p}_{t_\ell})U_{\ell}}_{\xi^{\mathrm{conc}}_\tau} + \underbrace{(\hat{p}_{t_\ell} - \bar{p}_{t_\ell})U_{\ell}}_{\xi^{\mathrm{bias}}_\tau} + \underbrace{\max_{j \in [J_{t_\ell}]} (\bar{p}_{t_\ell} - \tilde{p}^j_{\ell})U_{\ell}}_{\xi^{\mathrm{opt}}_\tau}.
\]
We bound the absolute values of these three terms separately (see Appendix~\ref{sec:app_b3_regret} for details):
\begin{itemize}[leftmargin=*]
    \item The empirical concentration term $\xi^{\mathrm{conc}}_\tau$ admits a variance-sensitive bound consisting of a leading square-root term of order $\sqrt{\Var_p(U_\ell)/n_\tau}$ together with a lower-order residual, while the pseudo-count bias term $\xi^{\mathrm{bias}}_\tau$ is controlled at the fast rate $\tilde{\mathcal O}(1/n_\tau)$ due to the identity $U_\ell(s_0)=0$.
    \item For the optimistic posterior term $\xi^{\mathrm{opt}}_\tau$, the Dirichlet concentration bound yields a pseudo-variance term of order $\mathcal O(\sqrt{\kappa_{t_\ell}\Var_{\bar p}(U_\ell)/n_\tau})$. A variance-sensitive perturbation argument together with a high/low variance case split (Appendix~\ref{sec:app_b3_regret}, Lemma~\ref{lem:variance_substitution}) converts this into the true variance without degrading the fast rate. Consequently,
    \[
    \sum_{\tau=1}^{T_K} |\mathrm{Error}_\tau|
    \le
    \tilde{\mathcal O}(1)\sqrt{SA\sum_{\tau=1}^{T_K}\Var_p(U_{\ell(\tau)})}
    + \tilde{\mathcal O}(B_\star S^2A).
    \]
\end{itemize}

\begin{figure}[!t]
    \centering
    \includegraphics[width=\textwidth]{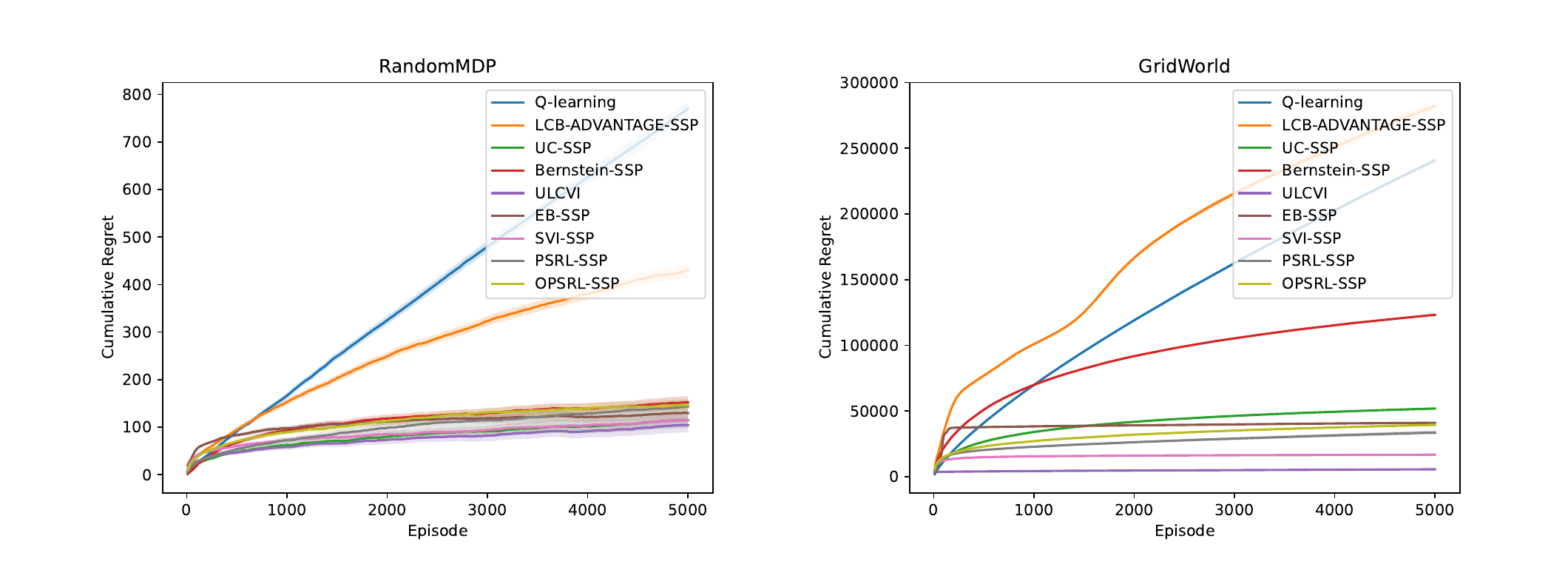}
    \caption{
        Cumulative regret of SSP algorithms on RandomMDP (left) and GridWorld (right) over 5,000 episodes. Curves are averaged over 500 runs; shaded areas show 95\% confidence intervals.
    }
    \label{fig:plot}
\end{figure}

\begin{figure}[!t]
    \centering
    \includegraphics[width=0.5\textwidth]{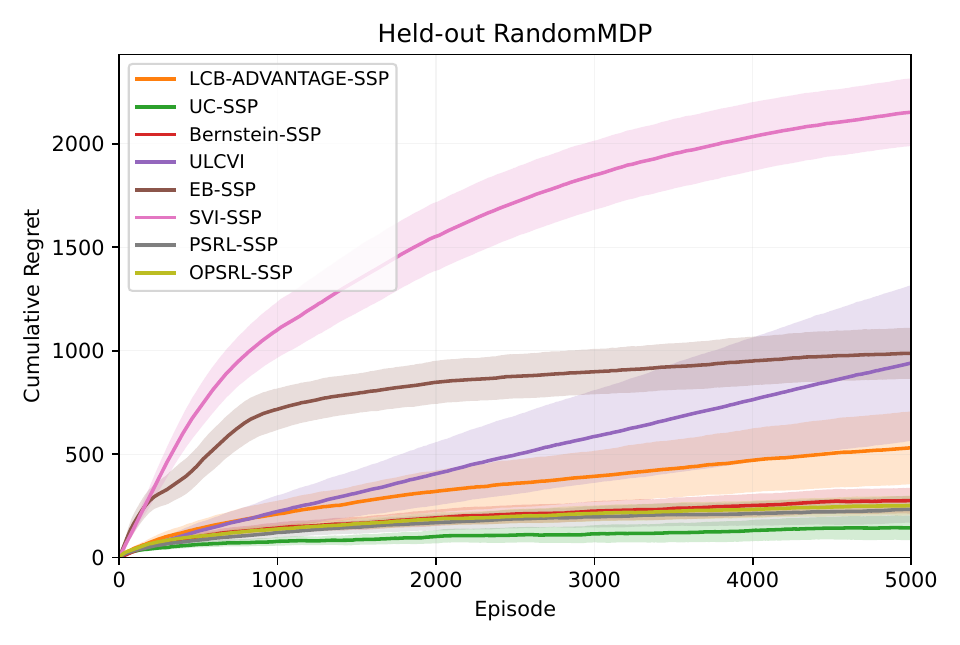}
    \caption{Cumulative regret on 20 independently generated held-out RandomMDP instances over 5,000 episodes. Curves average 20 rollouts within each instance; shaded areas show pointwise 95\% confidence intervals across instances.}
    \label{fig:heldout_randommdp}
\end{figure}

\paragraph{Step 3: Variance-Aware Analysis and Self-Bounding.}
To obtain a variance-sensitive bound for the martingale term, we apply Freedman's inequality to both the primary sequence $\sum \widetilde{M}_\tau$ and the squared-MDS sequence. This reduces the analysis to bounding the cumulative conditional variance, which satisfies along the true trajectory:
\[
\sum_{\tau=1}^{T_K} \Var_p(U_{\ell(\tau)}) \le \Otilde\left( B_\star^2 K + B_\star R_K + B_\star^2 SA \right) + \Otilde(B_\star) \sum_{\tau=1}^{T_K} |\text{Error}_\tau|.
\]
Substituting this variance bound into the regret decomposition yields the following self-bounding inequality for $R_K$ (see Appendix~\ref{sec:app_b3_regret}):
\[
R_K \le \Otilde\left( \sqrt{SA \cdot \left( B_\star^2 K + B_\star R_K + B_\star^2 S^2 A \right)} + B_\star S^2 A \right).
\]
Applying the AM-GM inequality and solving the resulting self-bounding inequality yields the regret bound in Theorem~1, with leading term $\Otilde(B_\star \sqrt{SAK})$ and lower-order term $\Otilde(B_\star S^2 A)$.

\section{Experiments}

In this section, we evaluate OPSRL-SSP in two complementary settings. The fixed RandomMDP and GridWorld benchmarks illustrate learning behavior on representative SSP instances, while independently generated held-out RandomMDP instances assess performance across different environments. Full experimental details are provided in Appendix~\ref{sec:appendix_e_experiments}.

\paragraph{Baselines} We compare OPSRL-SSP with PSRL-SSP \citep{jafarniajahromi2021onlinelearningstochasticshortest}, the OFU-based algorithms UC-SSP \citep{tarbouriech2020noregretexplorationgoalorientedreinforcement}, Bernstein-SSP \citep{cohen2020nearoptimalregretboundsstochastic}, ULCVI \citep{cohen2021minimaxregretstochasticshortest}, EB-SSP \citep{tarbouriech2021stochasticshortestpathminimax}, SVI-SSP, and LCB-ADVANTAGE-SSP \citep{chen2021implicitfinitehorizonapproximationefficient}, and the $\epsilon$-greedy Q-learning heuristic. Appendix~\ref{sec:appendix_e_experiments} provides the environment setups and hyperparameters for OPSRL-SSP and all baselines.

\paragraph{Results}
Figure~\ref{fig:plot} reports cumulative regret on the fixed RandomMDP and GridWorld instances. On these instances, OPSRL-SSP performs competitively with ULCVI, SVI-SSP, EB-SSP, and PSRL-SSP, and outperforms LCB-ADVANTAGE-SSP and $\epsilon$-greedy Q-learning. These results indicate that OPSRL-SSP retains the empirical efficiency of posterior sampling while performing comparably to the tuned OFU methods.

Figure~\ref{fig:heldout_randommdp} reports regret on 20 independently generated held-out RandomMDP instances. Each method has 20 rollouts per instance, and the confidence intervals use instances as the statistical units. Across these instances, OPSRL-SSP achieves lower mean endpoint cumulative regret and nAUC than the tested ULCVI, EB-SSP, and LCB-ADVANTAGE-SSP configurations, while remaining competitive with Bernstein-SSP and PSRL-SSP. Together with the fixed-instance results in Figure~\ref{fig:plot}, this evaluation shows that OPSRL-SSP retains competitive learning performance across multiple independently generated RandomMDP environments.

\FloatBarrier

\section{Conclusion}

In this paper, we introduced OPSRL-SSP, the first optimistic posterior-sampling algorithm with a regret bound whose leading term is minimax-optimal, $\tilde{\mathcal{O}}(B_\star \sqrt{SAK})$, for the stochastic shortest path (SSP) setting. By augmenting the state space with an optimistic pseudo-state and using dynamic posterior inflation, our approach establishes global anytime optimism under unbounded horizons and removes the additional $\sqrt{S}$ factor in the PSRL-SSP regret bound. We also developed a cost-perturbation analysis for zero-cost SSPs and showed how the regret guarantee for the perturbed problem transfers to the original problem.

Although our algorithm achieves a minimax-optimal leading regret term, it requires knowledge of $B_\star$. A natural direction for future work is to remove this requirement using a doubling scheme similar to that in~\cite{chen2021findingstochasticshortestpath}. Extending optimistic posterior sampling to broader SSP models, including large-scale settings with function approximation, is another direction for future research.

\section*{Acknowledgment}
This work was supported by Natural Science Foundation of China (No. 62306116).


\clearpage
\bibliographystyle{plainnat}
\bibliography{refs}

\clearpage
\appendix

\addtocontents{toc}{\protect\setcounter{tocdepth}{2}}

\noindent{\Huge \textbf{Appendix}}
\vspace{2.5em}

\noindent{\LARGE \textbf{Table of Contents}}
\vspace{0.5em}
\hrule height 1pt
\vspace{0.5em}

\makeatletter
\@starttoc{toc}
\makeatother

\vspace{1em}
\hrule height 0.5pt
\vspace{2em}

\clearpage 

\section{Notation and Preliminaries}
\label{sec:appendix_a_notation}

\begin{table}[H]
\centering
\caption{Notation used throughout the paper.}
\label{tab:notation}
\small
\setlength{\tabcolsep}{4pt}
\begin{tabular}{@{}ll@{}}
\toprule
\textbf{Notation} & \textbf{Meaning} \\
\midrule
$\mathcal{S}$ & state space of size $S$ \\
$\mathcal{A}$ & action space of size $A$ \\
$\mathcal{S}^+$ & physical state space including the goal state, $\mathcal{S}^+ = \mathcal{S} \cup \{g\}$ \\
$\tilde{\mathcal{S}}$ & internal augmented state space used for optimistic planning, $\tilde{\mathcal{S}} = \mathcal{S}^+ \cup \{s_0\}$ \\
$s_{init}$ & fixed initial state \\
$g$ & absorbing goal state \\
$L_M$ & total number of epochs \\
$K$ & number of episodes \\
$T_K$ & total number of steps in $K$ episodes \\
$J_t$ & dynamic number of posterior samples at global step $t$ \\
$B_\star$ & upper bound on the expected cost of the optimal policy \\
\midrule
$c(s,a)$ & instantaneous cost function \\
$c_{\min}$ & minimum instantaneous cost \\
$\epsilon$ & cost perturbation parameter (for zero-cost environments) \\
$p(\cdot|s, a)$ & true physical transition kernel mapping to $\Delta(\mathcal{S}^+)$ \\
$\tilde{p}_\ell(\cdot|s, a)$ & sampled internal transition kernel mapping to $\Delta(\tilde{\mathcal{S}})$ \\
$V^\pi(s_t)$ & V-function of a given policy $\pi$  \\
$Q^\star(s_t, a_t)$ & optimal Q-function \\
$V^\star(s_t)$ & optimal V-function on the physical state space $\mathcal{S}^+$ \\
$\bar{V}^\star(s)$ & extended optimal V-function on the augmented space $\tilde{\mathcal{S}}$ with $\bar{V}^\star(s_0) = -2B_\star$ \\
$R_K$ & cumulative regret over $K$ episodes \\
\midrule
$n_{0,t}(s,a)$ & dynamic initial pseudo-count \\
$s_0$ & optimistic pseudo-state \\
$c_0$ & constant used in pseudo-count calculation \\
$\kappa_t(s,a)$ & dynamic posterior inflation parameter \\
\midrule
$s_t$ & state that was visited at $t$-th step \\
$a_t$ & action that was picked at $t$-th step \\
$n_t(s, a)$ & number of visits of state-action $n_t(s, a) = \sum_{\tau=1}^{t-1} \mathbbm{1}\{(s_\tau, a_\tau) = (s, a)\}$ \\
$n_t(s'|s, a)$ & number of transitions to $s'$ from state-action $n_t(s'|s, a) = \sum_{\tau=1}^{t-1} \mathbbm{1}\{(s_\tau, a_\tau, s_{\tau+1}) = (s, a, s')\}$ \\
$\bar{n}_t(s, a)$ & extended number of visits $\bar{n}_t(s, a) = n_t(s, a) + n_{0,t}(s,a)$ \\
$\bar{n}_t(s'|s, a)$ & extended transitions to $s'$: $\bar{n}_t(s'|s, a) = n_t(s'|s, a) + \mathbbm{1}\{s' = s_0\} \cdot n_{0,t}(s,a)$ \\
$\widehat{p}_t(s'|s, a)$ & empirical probability transition $\widehat{p}_t(s'|s, a) = n_t(s'|s, a) / n_t(s, a)$ \\
$\bar{p}_t(s'|s, a)$ & pseudo-empirical probability transition $\bar{p}_t(s'|s, a) = \bar{n}_t(s'|s, a) / \bar{n}_t(s, a)$ \\
\midrule
$\tilde{Q}_\ell(s, a)$ & optimistic lower bound of the optimal Q-value \\
$\tilde{V}_\ell(s)$    & optimistic lower bound of the optimal V-value \\
$U_\ell(s)$ & shifted optimistic value function, defined as $U_\ell(s)=\tilde{V}_\ell(s)+2B_\star$ \\
\midrule
$\mathbb{R}_+$ & non-negative real numbers \\
$\mathbb{R}_{++}$ & positive real numbers \\
$\mathbb{N}_{++}$ & positive natural numbers \\
$[n]$ & set $\{1, 2, \ldots, n\}$ \\
$\Delta_d$ & $d$-dimensional probability simplex: $\Delta_d = \{x \in \mathbb{R}_+^{d+1} : \sum_{j=0}^d x_j = 1 \}$ \\
\bottomrule
\end{tabular}
\end{table}

Let $(\mathcal{X}, \mathscr{X})$ be a measurable space and $\mathcal{P}(\mathcal{X})$ be the set of all probability measures on this space. For $p \in \mathcal{P}(\mathcal{X})$ we denote by $\mathbb{E}_p$ the expectation w.r.t.\ $p$. For random variable $\xi : \mathcal{X} \to \mathbb{R}$ notation $\xi \sim p$ means $\text{Law}(\xi) = p$. We also write $\mathbb{E}_{\xi \sim p}$ instead of $\mathbb{E}_p$. For independent (resp.\ i.i.d.) random variables $\xi_\ell \stackrel{\text{ind}}{\sim} p_\ell$ (resp.\ $\xi_\ell \stackrel{\text{i.i.d.}}{\sim} p$), $\ell = 1, \ldots, d$, we will write $\mathbb{E}_{\xi_\ell \stackrel{\text{ind}}{\sim} p_\ell}$ (resp.\ $\mathbb{E}_{\xi_\ell \stackrel{\text{i.i.d.}}{\sim} p}$), to denote expectation w.r.t.\ product measure on $(\mathcal{X}^d, \mathscr{X}^{\otimes d})$. For any $p, q \in \mathcal{P}(\mathcal{X})$ the Kullback-Leibler divergence $\text{KL}(p, q)$ is given by
\begin{equation*}
    \text{KL}(p, q) \triangleq
    \begin{cases}
        \mathbb{E}_p \left[ \log \frac{\mathrm{d}p}{\mathrm{d}q} \right], & p \ll q, \\
        +\infty, & \text{otherwise}.
    \end{cases}
\end{equation*}
For any $p \in \mathcal{P}(\mathcal{X})$ and $f : \mathcal{X} \to \mathbb{R}$, $pf = \mathbb{E}_p[f]$. In particular, for any $p \in \Delta_d$ and $f : \{0, \ldots, d\} \to \mathbb{R}$, $pf = \sum_{\ell=0}^d f(\ell)p(\ell)$. Define $\Var_p(f) = \mathbb{E}_{s' \sim p}\left[ (f(s') - pf)^2 \right] = p[f^2] - (pf)^2$. For any $(s, a) \in \mathcal{S} \times \mathcal{A}$, true transition kernel $p(s, a) \in \mathcal{P}(\mathcal{S}^+)$ and $f : \mathcal{S}^+ \to \mathbb{R}$ define $pf(s, a) = \mathbb{E}_{p(s,a)}[f]$ and $\Var_p[f](s, a) = \Var_{p(s,a)}[f]$.

We write $f_1(S, A, B_\star, K) = \mathcal{O}(f_2(S, A, B_\star, K, \delta))$ if there exist $S_0, A_0, K_0, \delta_0$ and a constant $C$ such that for any $S \ge S_0$, $A \ge A_0$, $K \ge K_0$, $\delta < \delta_0$, and any valid expected cost bound $B_\star > 0$, $f_1(S, A, B_\star, K, \delta) \le C \cdot f_2(S, A, B_\star, K, \delta)$. We write $f_1(S, A, B_\star, K, \delta) = \tilde{\mathcal{O}}(f_2(S, A, B_\star, K, \delta))$ if the factor $C$ in the previous definition is poly-logarithmic in $S, A, B_\star, K, 1/\delta$.

For $k, \theta > 0$ define $\Gamma(k, \theta)$ as a gamma-distribution with a shape parameter $k$ and a rate parameter $\theta$.

We fix a non-negative cost function $f : \{1, \ldots, m\} \to [0, B_\star]$ and define the lower-tail minimum Kullback-Leibler divergence for $p \in \Delta_{m-1}$ and $u \in [0, B_\star]$:
\begin{equation*}
    \mathcal{K}_{\inf}^{\mathrm{lower}}(p, u, f) \triangleq \inf \left\{ \text{KL}(p, q) : q \in \Delta_{m-1}, qf \le u \right\}.
\end{equation*}

Similar to the standard Kullback-Leibler divergence, this lower-tail quantity admits a variational formula up to rescaling for any $u \in [0, B_\star]$:
\begin{equation*}
    \mathcal{K}_{\inf}^{\mathrm{lower}}(p, u, f) = \max_{\lambda \in \left[0, \frac{1}{u}\right]} \mathbb{E}_{X \sim p} \left[ \log(1 - \lambda(u - f(X))) \right].
\end{equation*}

\clearpage
\section{Detailed Proofs for OPSRL-SSP}
\label{sec:appendix_b_proofs}

\subsection{Concentration Events}
\label{sec:app_b1_concentration}

Let $\beta^\star, \beta^{KL}, \beta^{conc} : (0, 1) \times \mathbb{N} \to \mathbb{R}_+$, $\beta^{Dir} : (0, 1) \times \mathbb{N} \times \mathbb{N} \to \mathbb{R}_+$, and $\beta : (0, 1) \times \mathbb{N} \to \mathbb{R}_+$ be some functions defined later in Lemma \ref{lem:concentration_bounds}. We define the following favorable events for the SSP setting:

\begingroup
\setlength{\jot}{1pt}
\begin{align*}
\mathcal{E}^\star(\delta) &\triangleq \Bigg\{ \forall t \in \mathbb{N}, \forall (s, a) \in \mathcal{S} \times \mathcal{A} : \\*
&\quad\; \mathcal{K}_{\inf}^{\mathrm{lower}}(\hat{p}_t(s, a), p V^\star(s, a), V^\star) \le \frac{\beta^\star(\delta, n_t(s, a))}{n_t(s, a)} \Bigg\}, \\[0.4ex]
\mathcal{E}^{KL}(\delta) &\triangleq \Bigg\{ \forall t \in \mathbb{N}, \forall (s, a) \in \mathcal{S} \times \mathcal{A} : \\*
&\quad\; \text{KL}(\hat{p}_t(s, a) \| p(s, a)) \le \frac{\beta^{KL}(\delta, n_t(s, a))}{n_t(s, a)} \Bigg\}, \\[0.4ex]
\mathcal{E}^{conc}(\delta) &\triangleq \Bigg\{ \forall t \in \mathbb{N}, \forall (s, a) \in \mathcal{S} \times \mathcal{A} : \\*
&\quad\; \big|(\hat{p}_t - p)V^\star(s, a)\big| \le \sqrt{\frac{2\Var_p [V^\star](s, a) \beta^{conc}(\delta, n_t(s, a))}{n_t(s, a)}} \\*
&\qquad + \frac{3B_\star \beta^{conc}(\delta, n_t(s, a))}{n_t(s, a)} \Bigg\}, \\[0.4ex]
\mathcal{E}^{Dir}(\delta) &\triangleq \Bigg\{ \forall t \in \mathbb{N}, \forall (s, a) \in \mathcal{S} \times \mathcal{A}, \forall j \in [J_t] : \\*
&\quad\; \big|(\tilde{p}_t^j - \bar{p}_t)U_{\ell(t)}(s, a)\big| \le 2 \sqrt{\frac{\Var_{\bar{p}_t} [U_{\ell(t)}](s, a) \beta^{Dir}(\delta, t, J_t) \cdot \kappa_t(s, a)}{\bar{n}_t(s, a)}} \\*
&\qquad + \frac{9B_\star \beta^{Dir}(\delta, t, J_t) \cdot \kappa_t(s, a)}{\bar{n}_t(s, a)} \Bigg\}, \\[0.4ex]
\mathcal{E}^{Freedman}(\delta) &\triangleq \Bigg\{ \forall t \in \mathbb{N} : \\*
&\quad\; \sum_{\tau=1}^t \Big( U_{\ell(\tau)}(s_{\tau+1}) - pU_{\ell(\tau)}(s_\tau, a_\tau) \Big) \le \sqrt{2 \sum_{\tau=1}^t \Var_p [U_{\ell(\tau)}](s_\tau, a_\tau) \beta(\delta, t)} \\*
&\qquad + 3B_\star \beta(\delta, t) \Bigg\}, \\[0.4ex]
\mathcal{E}^{Freedman\_sq}(\delta) &\triangleq \Bigg\{ \forall t \in \mathbb{N} : \\*
&\quad\; \sum_{\tau=1}^t \Big( p[U^2_{\ell(\tau)}](s_\tau, a_\tau) - U^2_{\ell(\tau)}(s_{\tau+1}) \Big) \le \sqrt{2 \sum_{\tau=1}^t \Var_p [U^2_{\ell(\tau)}](s_\tau, a_\tau) \beta(\delta, t)} \\*
&\qquad + 9B_\star^2 \beta(\delta, t) \Bigg\}.
\end{align*}
\endgroup

We also introduce the intersection of these events, $\mathcal{G}^{conc}(\delta) \triangleq \mathcal{E}^\star(\delta) \cap \mathcal{E}^{KL}(\delta) \cap \mathcal{E}^{conc}(\delta) \cap \mathcal{E}^{Dir}(\delta) \cap \mathcal{E}^{Freedman}(\delta) \cap \mathcal{E}^{Freedman\_sq}(\delta)$.

\begin{lemma}\label{lem:concentration_bounds}
For any $\delta \in (0, 1)$ and for the following choices of functions $\beta$,
\begin{align*}
    \beta^\star(\delta, n) &\triangleq \log(12SA/\delta) + 3 \log(e\pi(2n + 1)), \\[0.5ex]
    \beta^{KL}(\delta, n) &\triangleq \log(12SA/\delta) + S \log(e(1 + n)), \\[0.5ex]
    \beta^{conc}(\delta, n) &\triangleq \log(12SA/\delta) + \log(4e(2n + 1)), \\[0.5ex]
    \beta^{Dir}(\delta, t, J_t) &\triangleq \log(24SA J_t t^2/\delta), \\[0.5ex]
    \beta(\delta, t) &\triangleq \log(48 t^2/\delta),
\end{align*}
it holds that the probability of each individual event is at least $1 - \delta/12$. In particular, $\mathbb{P}[\mathcal{G}^{conc}(\delta)] \ge 1 - \delta/2$.
\end{lemma}

\begin{remark}
Since $J_t = \Theta(\log(SA t/\delta))$, all functions $\beta$ are polylogarithmic in $S, A, B_\star, K, t$, and $1/\delta$. For $\beta^{Dir}$ and $\beta$, the logarithmic dependence on $t^2$ makes the bounds uniform over time, accounting for the fact that the total number of steps $T_K$ in SSP is a random stopping time.
\end{remark}

\begin{proof}[Proof of Lemma \ref{lem:concentration_bounds}]
$\mathbb{P}[\mathcal{E}^\star(\delta)] \ge 1 - \delta/12$ follows from \textbf{Lemma \ref{lem:lower_tail_kl_concentration}}. Applying \textbf{Lemma \ref{lem:kl_method_of_types}} and the union bound over $(s, a) \in \mathcal{S} \times \mathcal{A}$, we get $\mathbb{P}[\mathcal{E}^{KL}(\delta)] \ge 1 - \delta/12$ using the standard KL concentration via the method of types.

Next, for $\mathcal{E}^{Dir}(\delta)$, since the Dirichlet sampling is evaluated at each global time step $t$, we apply a standard union bound over $j \in [J_t]$, $(s, a) \in \mathcal{S} \times \mathcal{A}$, and all possible time steps $t \in \mathbb{N}$. Setting the failure probability at step $t$ to $\delta' = \frac{\delta}{12SAJ_t \cdot (\frac{\pi^2}{6} t^2)} \ge \frac{\delta}{24SAJ_t t^2}$, the sum of failure probabilities over $t \in \mathbb{N}$ is bounded by $\delta/12$. This yields the time-uniform term $\beta^{Dir}$, concluding $\mathbb{P}[\mathcal{E}^{Dir}(\delta)] \ge 1 - \delta/12$.

Applying the self-normalized Bernstein-type concentration inequality for empirical weighted sums, and then taking a union bound over $(s,a)$, yields $\mathbb{P}[\mathcal{E}^{conc}(\delta)] \ge 1-\delta/12$.

To bound $\mathbb{P}[\mathcal{E}^{Freedman}(\delta)]$ and $\mathbb{P}[\mathcal{E}^{Freedman\_sq}(\delta)]$, we consider the martingale difference sequences associated with the shifted optimistic value functions. The sequence $\widetilde{M}_\tau = U_{\ell(\tau)}(s_{\tau+1}) - pU_{\ell(\tau)}(s_\tau, a_\tau)$ has absolute value at most $3B_\star$. Applying the anytime variance-aware Freedman inequality (\textbf{Lemma \ref{lem:anytime_freedman}}) with a time-uniform union bound over $t \in \mathbb{N}$ (using the $\frac{\pi^2}{6}t^2$ weighting) gives $\mathbb{P}[\mathcal{E}^{Freedman}(\delta)] \ge 1 - \delta/12$. Similarly, the squared sequence $\widetilde{M}'_\tau = p[U^2_{\ell(\tau)}] - U^2_{\ell(\tau)}(s_{\tau+1})$ has absolute value at most $9B_\star^2$. Applying the anytime Freedman inequality to $\widetilde{M}'_\tau$ gives $\mathbb{P}[\mathcal{E}^{Freedman\_sq}(\delta)] \ge 1 - \delta/12$ with $\beta(\delta, t) = \log(48 t^2/\delta)$.
\end{proof}

\subsection{Optimism Guarantee for OPSRL-SSP}
\label{sec:app_b2_optimism}

\begin{proposition}[Global Anytime Optimism for OPSRL-SSP]
\label{prop:optimism_ssp}
Assume that at any global time step $t$, the number of posterior samples is $J_t = \lceil c_J \cdot \ln(4 S A \pi^2 t^2 / \delta) \rceil$ where $c_J \triangleq 1 / (-\ln(1-\gamma))$ and $\gamma > 0$ is the absolute constant lower bound for the single-sample optimism probability. The dynamic inflation parameter is $\kappa_t(s,a) = 2\beta^\star(\delta, n_t(s,a))$, and the dynamic initial pseudo-count is
\[
n_{0,t}(s, a)
=
\left\lceil
\kappa_t(s, a)\Big(c_0+\log_{17/16}(n_t(s, a)/\kappa_t(s, a)+1)\Big)
\right\rceil,
\]
where
$c_0 \triangleq c_0^{\mathrm{Dir}}(1/2)+1$
and $c_0^{\mathrm{Dir}}(1/2)$ denotes the absolute constant from \textbf{{Lemma \ref{lem:dirichlet_anti_concentration}}} with $\varepsilon=1/2$. We augment the state space with an optimistic pseudo-state $s_0$ and define the extended optimal value $\bar{V}^\star$ such that $\bar{V}^\star|_{\mathcal{S}^+} = V^\star$ and $\bar{V}^\star(s_0) \triangleq -2B_\star$. Then, conditional on the concentration event $\mathcal{E}^\star$, the anytime anti-concentration event
\begin{equation}
\mathcal{E}^{\mathrm{anticonc}}(\delta) \triangleq \left\{ \forall t \in \mathbb{N}, \forall (s, a) \in \mathcal{S} \times \mathcal{A} : \min_{j \in [J_t]} \left\{ \tilde{p}_t^j \bar{V}^\star(s, a) \right\} \le p V^\star(s, a) \right\}
\end{equation}
holds with probability at least $1 - \delta/2$.
\end{proposition}

\begin{proof}
For any fixed $t,j,s,a$, the sampled transition vector $\tilde{p}_t^j(\cdot|s,a)$ follows $\operatorname{Dir}((\bar{n}_t(s'|s,a)/\kappa_t(s,a))_{s'\in\tilde{\mathcal{S}}})$ on the augmented state space $\tilde{\mathcal{S}}=\mathcal{S}\cup\{g,s_0\}$. Some coordinates may have zero parameters under our improper-Dirichlet convention, so we consider the distribution on its effective support. Define
\[
\mathcal{I}_t(s,a)\triangleq \{s'\in\tilde{\mathcal S}:\bar n_t(s'|s,a)>0\}.
\]
Then $\tilde p_t^j(\cdot|s,a)$ is obtained by sampling a proper Dirichlet distribution on the lower-dimensional simplex over $\mathcal{I}_t(s,a)$ with parameter vector
\[
\left(\frac{\bar n_t(s'|s,a)}{\kappa_t(s,a)}\right)_{s'\in\mathcal{I}_t(s,a)},
\]
and setting all coordinates outside $\mathcal{I}_t(s,a)$ deterministically to zero. Therefore, whenever we invoke Lemma~\ref{lem:dirichlet_anti_concentration}, it is applied to this proper Dirichlet law on the effective support $\mathcal{I}_t(s,a)$.

To apply the Gaussian lower bound for Dirichlet distributions (\textbf{Lemma \ref{lem:dirichlet_anti_concentration}}), which is designed for upper-tail probabilities (reward maximization), we define the extended optimal value function $\bar{V}^\star: \tilde{\mathcal{S}} \to \mathbb{R}$ such that $\bar{V}^\star|_{\mathcal{S}^+} = V^\star$ and $\bar{V}^\star(s_0) \triangleq -2B_\star$. We then introduce an auxiliary function $f(s) = B_\star - \bar{V}^\star(s)$.
For each $s\in\mathcal{S}^+$, $V^\star(s) \in [0, B_\star]$ (with $V^\star(g)=0$), so $f(s) \in [0, B_\star]$, and we take $b = B_\star$. At the pseudo-state, $f(s_0) = B_\star - (-2B_\star) = 3B_\star$. This satisfies the condition $b_0 = 3B_\star \ge 2b$ required by \textbf{Lemma \ref{lem:dirichlet_anti_concentration}}.

Let us define
$\alpha_0 = \frac{n_{0,t}(s,a)}{\kappa_t(s,a)} - 1$ and $\alpha_i = \frac{n_t(s_i|s,a)}{\kappa_t(s,a)}
\quad \text{for } s_i \in \mathcal{S}^+.$
Then the Dirichlet parameter vector takes the form
$(\alpha_0+1,\alpha_1,\dots,\alpha_{S+1}),$
and its total mass is
$\bar{\alpha}
=
(\alpha_0+1)+\sum_i \alpha_i
=
\frac{\bar{n}_t(s,a)}{\kappa_t(s,a)}.$
Let $q \in \Delta_{S+1}$ denote the corresponding mean distribution, i.e.
$q(s_0)=\frac{\alpha_0+1}{\bar{\alpha}},$
and
$q(s_i)=\frac{\alpha_i}{\bar{\alpha}}
\quad \text{for } s_i\in\mathcal{S}^+.$
By our dynamic choice of $n_{0,t}(s,a)$, we have
\[
\alpha_0 + 1
=
\frac{n_{0,t}(s,a)}{\kappa_t(s,a)}
\ge
c_0 + \log_{17/16}\!\left(\frac{n_t(s,a)}{\kappa_t(s,a)} + 1\right).
\]
Hence, the pseudo-state parameter grows at least logarithmically with the empirical scale $n_t(s,a)/\kappa_t(s,a)$. Moreover, since
\[
\bar{\alpha}
=
\frac{n_t(s,a)}{\kappa_t(s,a)} + (\alpha_0+1),
\]
enlarging the absolute constant $c_0$ if necessary ensures that the parameter-growth requirement in \textbf{Lemma~\ref{lem:dirichlet_anti_concentration}} is satisfied.

\paragraph{Case 1: $\bar{\alpha} \ge 2\alpha_0$.}
By applying \textbf{Lemma \ref{lem:dirichlet_anti_concentration}} to the auxiliary function $f$, for any target $u' \ge qf$, we obtain the Gaussian lower bound:

\begin{equation}
\mathbb{P} \left( \tilde{p}_t^j f \ge u' \right) \ge \frac{1}{2} \left( 1 - \Phi \left( \sqrt{\frac{2\bar{n}_t(s,a)}{\kappa_t(s,a)} \mathcal{K}_{\inf}(q, u', f)} \right) \right),
\label{eq:gaussian_bound_f}
\end{equation}

where $\Phi$ is the CDF of the standard normal distribution. Set $u'=B_\star-u$. The event $\tilde{p}_t^j f \ge B_\star - u$ is equivalent to $\tilde{p}_t^j \bar{V}^\star \le u$. By the definition of the minimum KL divergence, $\mathcal{K}_{\inf}(q, B_\star - u, B_\star - \bar{V}^\star) = \mathcal{K}_{\inf}^{\mathrm{lower}}(q, u, \bar{V}^\star)$.
If $u' < qf$ (which implies $u > q\bar{V}^\star$), the probability is at least $\frac{1}{2}(1 - \Phi(0)) = 0.25$. Since $\mathcal{K}_{\inf}^{\mathrm{lower}}(q, u, \bar{V}^\star) = 0$ in this regime, the inequality holds for all $u \ge 0$.

\paragraph{Case 2: $\bar{\alpha} < 2\alpha_0$.}
In this regime, the pseudo-state heavily dominates the distribution. Since $(\alpha_0+1) / \bar{\alpha} > 1/2$ and $f(s_0) = 3B_\star$, we have $qf > 1.5B_\star \ge B_\star$. For any valid expected cost target $u \in [0, B_\star]$, the corresponding $u' = B_\star - u \le B_\star < qf$. We can lower-bound the probability by considering the marginal Beta distribution $\xi \sim \operatorname{Beta}(\alpha_0 + 1, \bar{\alpha} - \alpha_0 - 1)$ representing the mass on $s_0$:
\begin{equation}
    \mathbb{P} \left( \tilde{p}_t^j f \ge u' \right) \ge \mathbb{P}_\xi \left( 3B_\star \xi \ge u' \right) \ge \mathbb{P}_\xi \left( \xi \ge \frac{1}{3} \right).
\end{equation}
Applying the Beta-distribution result of \citet{alfers1984normal} and using $(\alpha_0+1)/\bar{\alpha} > 1/2$, we obtain a lower bound strictly greater than $\frac{1}{2}(1 - \Phi(0))$. Thus, the same lower bound applies in this case.

\paragraph{Establishing the Constant Success Rate.}
We now apply the translated lower bound to the true environment's expected cost $u = p V^\star(s, a) \le B_\star$. Since the true transition $p$ is supported on $\mathcal{S}^+$, $p\bar{V}^\star = pV^\star$:
\begin{equation}
    \mathbb{P} \left( \tilde{p}_t^j \bar{V}^\star(s, a) \le p V^\star(s, a) \right) \ge \frac{1}{2} \left( 1 - \Phi \left( \sqrt{\frac{2\bar{n}_t(s, a)}{\kappa_t(s,a)} \mathcal{K}_{\inf}^{\mathrm{lower}}(q, p V^\star, \bar{V}^\star)} \right) \right).
    \label{eq:lower_tail_bound}
\end{equation}
Using the variational formula for $\mathcal{K}_{\inf}^{\mathrm{lower}}$ with $\lambda$ rescaled to $[0, 1]$, and noting the maximum deviation is $u - \bar{V}^\star(s_0) = u + 2B_\star$, we decouple the pseudo-count from the empirical transitions:
\begin{align}
    \bar{n}_t(s, a) \mathcal{K}_{\inf}^{\mathrm{lower}}(q, u, \bar{V}^\star)
    &= \bar{n}_t(s, a) \max_{\lambda \in [0,1]} \mathbb{E}_{s' \sim q} \left[ \log \left( 1 - \lambda \frac{u - \bar{V}^\star(s')}{2B_\star + u} \right) \right] \nonumber \\*
    &\le \max_{\lambda \in [0,1]} \bigg\{ n_{0,t}(s,a) \log(1 - \lambda) \nonumber \\*
    &\qquad\quad + (\bar{n}_t(s, a) - n_{0,t}(s,a)) \mathbb{E}_{s' \sim \hat{p}_t} \left[ \log \left( 1 - \lambda \frac{u - \bar{V}^\star(s')}{2B_\star + u} \right) \right] \bigg\} \nonumber \\*
    &\le n_t(s, a) \max_{\lambda \in [0,1]} \mathbb{E}_{s' \sim \hat{p}_t} \left[ \log \left( 1 - \lambda \frac{u - \bar{V}^\star(s')}{2B_\star + u} \right) \right] \nonumber \\*
    &\le n_t(s, a) \max_{\lambda' \in [0,1]} \mathbb{E}_{s' \sim \hat{p}_t} \left[ \log \left( 1 - \lambda' \frac{u - V^\star(s')}{u} \right) \right] \nonumber \\*
    &= n_t(s, a) \mathcal{K}_{\inf}^{\mathrm{lower}}(\hat{p}_t(s, a), u, V^\star).
\end{align}
Here, the first inequality isolates the pseudo-state contribution and the mass of the true empirical distribution. The second inequality drops the non-positive term $n_{0,t}(s,a)\log(1-\lambda)$, since $\log(1-\lambda)\le 0$ for all $\lambda\in[0,1]$.
The final inequality uses the fact that $\hat{p}_t$ assigns mass only to the physical states $\mathcal{S}^+$, where $\bar{V}^\star(s') \equiv V^\star(s')$.

If $u>0$, we perform the reparameterization
\[
\lambda' = \lambda \frac{u}{2B_\star + u} \in [0,1].
\]
Because $\hat{p}_t$ is supported on $\mathcal{S}^+$, where the minimum possible value is $V^\star(g)=0$, this yields
\[
\bar{n}_t(s, a) \mathcal{K}_{\inf}^{\mathrm{lower}}(q, u, \bar{V}^\star)
\le
n_t(s, a) \mathcal{K}_{\inf}^{\mathrm{lower}}(\hat{p}_t(s, a), u, V^\star).
\]

If $u=0$, we bypass the above reparameterization and work directly with the auxiliary upper-tail formulation $f_\star = B_\star - \bar{V}^\star$. In this case, the preceding display implies
\[
\bar n_t(s,a)\,\mathcal K_{\inf}^{\mathrm{lower}}(q,0,\bar V^\star)
\le
n_t(s,a)\,\mathcal K_{\inf}(\hat p_t(s,a),B_\star,B_\star-V^\star)
=
n_t(s,a)\,\mathcal K_{\inf}^{\mathrm{lower}}(\hat p_t(s,a),0,V^\star).
\]
Hence, the same comparison bound holds for all $u \in [0,B_\star]$.

On the concentration event $\mathcal{E}^\star$, the lower-tail KL divergence of the empirical transition row $\hat{p}_t$ is bounded by the confidence budget $\beta^\star$:
\begin{equation}
    n_t(s, a) \mathcal{K}_{\inf}^{\mathrm{lower}}(\hat{p}_t(s, a), p V^\star(s, a), V^\star) \le \beta^\star(\delta, n_t(s, a)).
\end{equation}
Substituting this bound into equation \eqref{eq:lower_tail_bound} and using the dynamic inflation parameter $\kappa_t(s,a) = 2\beta^\star(\delta, n_t(s, a))$ gives:
\begin{equation}
    \mathbb{P} \left( \tilde{p}_t^j \bar{V}^\star(s, a) \le p V^\star(s, a) \mid \mathcal{E}^\star(\delta) \right) \ge \frac{1}{2} \left( 1 - \Phi(1) \right) \triangleq \gamma.
\end{equation}
This gives a uniform lower bound $\gamma=\frac{1}{2}(1-\Phi(1))\approx0.079$ on the probability of an optimistic sample at every time step, independently of the random and potentially unbounded episode lengths in SSP.

\paragraph{Time-Uniform Union Bound.}
Finally, taking the minimum over $J_t = \lceil c_J \cdot \ln(4 S A \pi^2 t^2 / \delta) \rceil = \lceil \frac{\ln(4 S A \pi^2 t^2 / \delta)}{-\ln(1-\gamma)} \rceil$ independent posterior samples, the probability of failure at a single state-action pair at global time step $t$ is bounded by:
\begin{equation}
    \mathbb{P} \left( \min_{j \in [J_t]} \left\{ \tilde{p}_t^j \bar{V}^\star(s, a) \right\} > p V^\star(s, a) \mid \mathcal{E}^\star(\delta) \right) \le (1 - \gamma)^{J_t} \le \frac{\delta}{4 S A \pi^2 t^2}.
\end{equation}
Taking a union bound over state-action pairs $(s, a) \in \mathcal{S} \times \mathcal{A}$ and summing over $t\in\mathbb{N}$ bounds the total failure probability by $\delta/24 \le \delta/2$. This establishes the global anytime anti-concentration event $\mathcal{E}^{\mathrm{anticonc}}(\delta)$.
\end{proof}

\begin{lemma}[Global Anytime Optimism via Strict Contraction]
\label{lem:optimism}
Let $\mathcal{E}^{\mathrm{opt}}$ be the global optimistic event such that the internal value function computed by Algorithm \ref{alg:opsrl_ssp} satisfies $\tilde{V}_\ell(s) \le V^\star(s)$ for all $\ell \ge 1$ and all physical states $s \in \mathcal{S}$. Conditioned on the time-uniform high-probability anti-concentration event $\mathcal{E}^{\mathrm{anticonc}}$, we have that $\mathcal{E}^{\mathrm{opt}}$ holds.
\end{lemma}

\begin{proof}
To extend one-step optimism to the value function in the unbounded-horizon SSP setting, we show that the injected pseudo-counts make the epoch-wise Bellman operator a strict contraction. Since $\mathcal{E}^{\mathrm{anticonc}}$ holds uniformly over global time steps, it also holds at any random epoch start $t_\ell$. We therefore fix an arbitrary epoch $\ell\ge1$ below.

\textbf{Step 1: Strict Contraction of the Augmented Bellman Operator.} \\
For any function $u: \mathcal{S} \to \mathbb{R}$, we define its fixed boundary extension $\bar{u}: \tilde{\mathcal{S}} \to \mathbb{R}$ such that $\bar{u}|_{\mathcal{S}} = u$, $\bar{u}(g) = 0$, and $\bar{u}(s_0) = -2B_\star$.
Let $T_\ell$ be the stationary optimistic Bellman operator defined for all $s \in \mathcal{S}$ as:
\begin{equation}
    (T_\ell u)(s) = \min_{a\in\mathcal{A}} \left[ c(s,a) + \min_{j\in[J_{t_\ell}]} \sum_{x \in \tilde{\mathcal{S}}} \tilde{p}_\ell^j(x|s,a)\bar{u}(x) \right].
\end{equation}
We first analyze the un-minimized internal mapping $F_{s,a,j}(u) = c(s,a) + \sum_{x \in \tilde{\mathcal{S}}} \tilde{p}_\ell^j(x|s,a)\bar{u}(x)$.
Because the boundary values for $g$ and $s_0$ are constants independent of $u$ and $v$, their differences are zero. Thus, for any two functions $u$ and $v$:
\begin{equation}
    |F_{s,a,j}(u) - F_{s,a,j}(v)| = \left| \sum_{x \in \mathcal{S}} \tilde{p}_\ell^j(x|s,a) \big(u(x) - v(x)\big) \right| \le \left( \sum_{x \in \mathcal{S}} \tilde{p}_\ell^j(x|s,a) \right) \|u - v\|_\infty.
\end{equation}
Although the prior may assign zero initial counts to physical states, the algorithm adds a positive dynamic pseudo-count $n_{0,t_\ell}(s,a)>0$ to the $s_0$ coordinate for every state-action pair. Thus, every sampled posterior model assigns positive probability to the pseudo-state almost surely. Since there are finitely many tuples $(s,a,j)$, define the minimum such mass as:
\begin{equation}
    \eta_\ell \triangleq \min_{s\in\mathcal{S}, a\in\mathcal{A}, j\in[J_{t_\ell}]} \tilde{p}_\ell^j(s_0|s,a) > 0 \quad \text{a.s.}
\end{equation}
Consequently, the probability mass on the active state space $\mathcal{S}$ is bounded above by $1-\eta_\ell$:
$\sum_{x \in \mathcal{S}} \tilde{p}_\ell^j(x|s,a) \le 1 - \tilde{p}_\ell^j(g|s,a) - \tilde{p}_\ell^j(s_0|s,a) \le 1 - \eta_\ell$.
This yields $|F_{s,a,j}(u) - F_{s,a,j}(v)| \le (1 - \eta_\ell) \|u - v\|_\infty$.

Since the minimum operators are 1-Lipschitz (i.e., $|\min_i \alpha_i - \min_i \beta_i| \le \max_i |\alpha_i - \beta_i|$), taking the minimum over actions $a$ and samples $j$ preserves the contraction rate:
\begin{equation}
    \|T_\ell u - T_\ell v\|_\infty \le \max_{s,a,j} |F_{s,a,j}(u) - F_{s,a,j}(v)| \le (1 - \eta_\ell) \|u - v\|_\infty.
\end{equation}
Because $1 - \eta_\ell < 1$, $T_\ell$ is a strict contraction on $(\mathbb{R}^{\mathcal{S}}, \|\cdot\|_\infty)$. By the Banach Fixed-Point Theorem, $T_\ell$ admits a \textbf{unique fixed point} $\tilde{V}_\ell$. This is the value function computed by the algorithm's internal value iteration in epoch $\ell$, so a separate proof of standard SSP properness for the augmented sampled models is not needed.

\textbf{Step 2: Single-Step Optimism under $\mathcal{E}^{\mathrm{anticonc}}$.} \\
Recall the true optimal value function $V^\star$ on $\mathcal{S}^+$, and define its extension $\bar{V}^\star$ such that $\bar{V}^\star|_{\mathcal{S}^+} = V^\star$ and $\bar{V}^\star(s_0) \triangleq -2B_\star$. Let $V^\star|_{\mathcal{S}}$ denote the restriction of $V^\star$ to $\mathcal{S}$, which serves as a valid input to the operator $T_\ell$.
On the anytime anti-concentration event $\mathcal{E}^{\mathrm{anticonc}}(\delta)$, for every $(s,a)$, the sampled optimistic models satisfy $\min_j \tilde{p}_\ell^j \bar{V}^\star(s,a) \le p V^\star(s,a)$. Evaluating $T_\ell$ at $V^\star|_{\mathcal{S}}$ gives:
\begin{align}
    (T_\ell (V^\star|_{\mathcal{S}}))(s)
    &= \min_{a \in \mathcal{A}} \left[ c(s, a) + \min_j \sum_{x \in \tilde{\mathcal{S}}} \tilde{p}_\ell^j(x|s,a)\bar{V}^\star(x) \right] \nonumber \\
    &\le \min_{a \in \mathcal{A}} \left[ c(s, a) + p V^\star(s, a) \right] = V^\star(s).
\end{align}
Thus, $T_\ell (V^\star|_{\mathcal{S}}) \le V^\star|_{\mathcal{S}}$ pointwise on $\mathcal{S}$.

\textbf{Step 3: Comparison Principle and Global Optimism.} \\
First, we observe that the operator $T_\ell$ is monotonically non-decreasing. For any $u \le v$ pointwise on $\mathcal{S}$, their fixed boundary extensions satisfy $\bar{u} \le \bar{v}$ pointwise on $\tilde{\mathcal{S}}$. Since the transition probabilities are non-negative, $\sum_x \tilde{p}_\ell^j \bar{u} \le \sum_x \tilde{p}_\ell^j \bar{v}$. Taking the minimum over $j$ and $a$ preserves this inequality, yielding $T_\ell u \le T_\ell v$.

Consider the sequence of functions generated by iteratively applying the operator starting from $u_0 = V^\star|_{\mathcal{S}}$, such that $u_{m+1} = T_\ell u_m$.
Since $u_1 = T_\ell (V^\star|_{\mathcal{S}}) \le V^\star|_{\mathcal{S}} = u_0$ (from Step 2), monotonicity guarantees that the sequence $\{u_m\}_{m=0}^\infty$ is monotonically decreasing:
\begin{equation}
    u_0 \ge u_1 \ge u_2 \ge \dots \ge u_m.
\end{equation}
Simultaneously, the strict contraction property (from Step 1) ensures that the sequence converges to the unique fixed point: $u_m \downarrow \tilde{V}_\ell$ as $m \to \infty$.
Since every element in this decreasing sequence is bounded above by $u_0 = V^\star|_{\mathcal{S}}$, its pointwise limit preserves the inequality:
\begin{equation}
    \tilde{V}_\ell(s) \le V^\star(s) \quad \forall s \in \mathcal{S}.
\end{equation}
Because this fixed-epoch argument holds uniformly for any $\ell \ge 1$ under $\mathcal{E}^{\mathrm{anticonc}}$, the global optimistic event $\mathcal{E}^{\mathrm{opt}}$ holds. This concludes the proof.
\end{proof}

\begin{lemma}[Bounded Range of Optimistic Value Function]
\label{lem:value_lower_bound}
The internal optimistic value function computed by Algorithm~\ref{alg:opsrl_ssp} is bounded from below. Specifically, for all epochs $\ell\ge1$ and states $s\in\tilde{\mathcal{S}}$:
\begin{equation}
    -2B_\star \le \tilde{V}_\ell(s) \le B_\star.
\end{equation}
\end{lemma}
\begin{proof}
The upper bound $\tilde{V}_\ell(s) \le V^\star(s) \le B_\star$ for $s \in \mathcal{S}$ follows from Lemma \ref{lem:optimism}, while $\tilde{V}_\ell(s_0) = -2B_\star \le B_\star$ holds by definition.
To prove the lower bound, consider the constant function $w(s) \equiv -2B_\star$ defined on $\mathcal{S}$, with boundary extensions $\bar{w}(g) = 0$ and $\bar{w}(s_0) = -2B_\star$.
Evaluating the operator $T_\ell$ on $w$ yields:
\begin{align}
    (T_\ell w)(s) &= \min_{a\in\mathcal{A}} \left[ c(s,a) + \min_{j\in[J_{t_\ell}]} \sum_{x \in \tilde{\mathcal{S}}} \tilde{p}_\ell^j(x|s,a)\bar{w}(x) \right] \nonumber \\
    &\ge \min_{a\in\mathcal{A}} \left[ 0 + \min_{j} \left( \tilde{p}_\ell^j(\mathcal{S}|s,a)(-2B_\star) + \tilde{p}_\ell^j(g|s,a)(0) + \tilde{p}_\ell^j(s_0|s,a)(-2B_\star) \right) \right] \nonumber \\
    &\ge -2B_\star = w(s)
\end{align}
where the first inequality uses $c(s,a) \ge 0$, and the second inequality holds because the sum of probabilities is 1 and $-2B_\star \le 0$. Thus, $T_\ell w \ge w$.
By the monotonicity and strict contraction of $T_\ell$ (established in Lemma \ref{lem:optimism}), iteratively applying $T_\ell$ starting from $w$ generates a monotonically increasing sequence that converges to the unique fixed point $\tilde{V}_\ell$. Therefore, $\tilde{V}_\ell(s) \ge w(s) = -2B_\star$ for all $s \in \mathcal{S}$. Combined with the boundary conditions, the bound holds globally on $\tilde{\mathcal{S}}$.
\end{proof}

\begin{remark}[Algebraic Invariance via Shifted Value Variable]
\label{remark:shifted_variable}
Because $\tilde{V}_\ell$ can take values as low as $-2B_\star$ due to the pseudo-state, standard concentration inequalities for nonnegative functions cannot be applied directly. To address this, we introduce the shifted value function for the remainder of the regret analysis:
\begin{equation}
    U_\ell(x) \triangleq \tilde{V}_\ell(x) + 2B_\star, \quad \forall x \in \tilde{\mathcal{S}}.
\end{equation}
By Lemma~\ref{lem:value_lower_bound}, $U_\ell$ is nonnegative and bounded: $0 \le U_\ell(x) \le 3B_\star$. For any two probability distributions $P$ and $Q$ on $\tilde{\mathcal{S}}$ (since $\sum P(x) = \sum Q(x) = 1$), the constant shift $2B_\star$ is canceled in expectation and variance differences:
\begin{equation}
    (P - Q)\tilde{V}_\ell = (P - Q)U_\ell, \qquad \Var_P(\tilde{V}_\ell) = \Var_P(U_\ell).
\end{equation}
Thus, we can use the shifted variable $U_\ell \in [0, 3B_\star]$ to analyze the estimation errors and variances while preserving the algebraic form of the Bellman residuals. The boundary condition gives $U_\ell(s_0)=0$, which is used to control the pseudo-count bias. At the goal state, however, $U_\ell(g)=2B_\star$. Therefore, telescoping arguments over episodes must account for this nonzero terminal value.
\end{remark}

\subsection{Derivation of the Minimax Optimal Regret Bound}
\label{sec:app_b3_regret}

In this section, we derive the regret bound of the OPSRL-SSP algorithm under the strictly positive cost assumption. In the following analysis, we assume that the high-probability joint concentration event $\mathcal{G}^{conc}(\delta)$ holds.

\begin{definition}[High-Probability Regret]
For a learning process consisting of $K$ episodes with a total of $T_K$ interaction steps, the high-probability regret is defined as:
\begin{equation}
    R_K = \sum_{k=1}^K \sum_{\tau=t_k}^{t_{k+1}-1} c(s_\tau, a_\tau) - K \cdot V^\star(s_{init}) \label{eq:regret_def}.
\end{equation}
\end{definition}

\begin{remark}[Measure-Theoretic Validity of Random Stopping Time]
In the SSP setting, $T_K$ is a random stopping time, and the states along the true physical trajectory satisfy $s_\tau \in \mathcal{S}$. Because the concentration inequalities used below hold uniformly over time, they also apply at the random stopping time $T_K$.
\end{remark}

\begin{lemma}[Number of epochs under the dual-stopping rule]
\label{lem:number_epochs}
Before the completion of \(K\) episodes and after \(T_K\) interaction steps, the total number of epochs satisfies
\[
L_M \le 1+SA+SA\log_2(T_K+1)+\sqrt{2K\left(1+SA+SA\log_2(T_K+1)\right)}.
\]
In particular,
\[
L_M=\tilde{\mathcal O}(\sqrt{SAK}+SA).
\]
\end{lemma}

\begin{proof}
We follow the standard macro-epoch argument of \citet{jafarniajahromi2021onlinelearningstochasticshortest}. Define a \emph{macro epoch} \(i\) as a consecutive block of epochs that starts whenever the count-doubling criterion is triggered. Let \(t_{u_1}=t_1\), and for \(i\ge 1\), define
\[
t_{u_{i+1}}=\min\{t_\ell>t_{u_i}: n_{t_\ell}(s,a)>2n_{t_{\ell-1}}(s,a)\text{ for some }(s,a)\in\mathcal S\times\mathcal A\}.
\]
Let \(N_M\) denote the total number of macro epochs up to time \(T_K\), and define \(u_{N_M+1}:=L_M+1\).

For epoch \(\ell\), recall that \(K_\ell\) denotes the number of completed episodes in that epoch. For macro epoch \(i\), define
\[
\widetilde K_i:=\sum_{\ell=u_i}^{u_{i+1}-1}K_\ell,
\]
the total number of completed episodes in macro epoch \(i\).

By construction of macro epochs, all epochs inside macro epoch \(i\), except possibly the last one, are terminated by the episode-increment criterion. Therefore,
\[
K_\ell \ge K_{\ell-1}+1, \qquad \ell=u_i,\dots,u_{i+1}-2.
\]
Hence, within macro epoch \(i\), the sequence of episode counts grows at least linearly, and we obtain
\[
\widetilde K_i=\sum_{\ell=u_i}^{u_{i+1}-1}K_\ell\ge \sum_{j=1}^{u_{i+1}-u_i-1}j=\frac{(u_{i+1}-u_i-1)(u_{i+1}-u_i)}{2}.
\]
Solving this quadratic inequality for \(u_{i+1}-u_i\) yields
\[
u_{i+1}-u_i\le 1+\sqrt{2\widetilde K_i}.
\]

Summing over all macro epochs, we get
\[
L_M=u_{N_M+1}-1=\sum_{i=1}^{N_M}(u_{i+1}-u_i)\le \sum_{i=1}^{N_M}(1+\sqrt{2\widetilde K_i})=N_M+\sum_{i=1}^{N_M}\sqrt{2\widetilde K_i}.
\]
Applying Cauchy--Schwarz,
\[
L_M\le N_M+\sqrt{2N_M\sum_{i=1}^{N_M}\widetilde K_i}.
\]
Since the total number of completed episodes across all macro epochs is exactly \(K\), we have
\[
\sum_{i=1}^{N_M}\widetilde K_i=K.
\]
Therefore,
\[
L_M\le N_M+\sqrt{2N_MK}.
\]

It remains to bound \(N_M\). A new macro epoch starts only when the count-doubling criterion is triggered. For each state-action pair \((s,a)\), let
\[
\mathcal T_{s,a}:=\{t_\ell\le T_K: n_{t_\ell}(s,a)>2n_{t_{\ell-1}}(s,a)\}
\]
be the set of epoch start times at which the doubling criterion is triggered by \((s,a)\). We claim that
\[
|\mathcal T_{s,a}|\le \lceil \log_2(n_{T_K+1}(s,a)+1)\rceil.
\]
If \(n_{T_K+1}(s,a)=0\), then \(|\mathcal T_{s,a}|=0\) and the claim is trivial. Assume now that \(n_{T_K+1}(s,a)\ge 1\). The first trigger associated with \((s,a)\) may occur when the count becomes positive from zero. Every subsequent trigger must occur at a time when the previous epoch-start count is already at least one, and each such trigger strictly more than doubles the count. Therefore, after \(m:=|\mathcal T_{s,a}|-1\) subsequent triggers, we must have
\[
n_{T_K+1}(s,a)>2^m.
\]
Hence,
\[
m<\log_2 n_{T_K+1}(s,a),
\]
which implies
\[
|\mathcal T_{s,a}| \le 1+\log_2 n_{T_K+1}(s,a)\le \lceil \log_2(n_{T_K+1}(s,a)+1)\rceil.
\]

Summing over all state-action pairs, we obtain
\[
N_M=1+\sum_{s,a}|\mathcal T_{s,a}|\le 1+\sum_{s,a}\lceil \log_2(n_{T_K+1}(s,a)+1)\rceil.
\]
Using \(\lceil x\rceil\le 1+x\), we get
\[
N_M\le 1+SA+\sum_{s,a}\log_2(n_{T_K+1}(s,a)+1).
\]
Since \(n_{T_K+1}(s,a)\le T_K\) for every \((s,a)\), it follows that
\[
N_M\le 1+SA+SA\log_2(T_K+1).
\]

Substituting this bound into \(L_M\le N_M+\sqrt{2N_MK}\) yields
\[
L_M\le 1+SA+SA\log_2(T_K+1)+\sqrt{2K\left(1+SA+SA\log_2(T_K+1)\right)}.
\]
This proves the stated bound. The \(\tilde{\mathcal O}(\sqrt{SAK}+SA)\) form follows immediately.
\end{proof}

\begin{lemma}[Regret Decomposition and Telescoping over Epochs]
\label{lem:regret_decomp}
Using the shifted variable $U_\ell = \tilde{V}_\ell + 2B_\star \in [0, 3B_\star]$, the regret can be bounded by the sum of single-step model estimation errors and cross-epoch truncation errors:
\begin{equation}
    R_K \le \sum_{\tau=1}^{T_K} \Big( U_{\ell(\tau)}(s_{\tau+1}) - \min_j \tilde{p}_{\ell(\tau)}^j U_{\ell(\tau)} \Big) + R_{trunc}
\end{equation}
where the truncation error satisfies $R_{\mathrm{trunc}}=\tilde{\mathcal O}(B_\star\sqrt{SAK}+B_\star SA)$.
\end{lemma}
\begin{proof}
By Lemma \ref{lem:optimism}, $\tilde{V}_{\ell(t_k)}(s_{init}) \le V^\star(s_{init})$. Thus, we can bound $R_K$ as follows:
\begin{equation}
    R_K \le \sum_{k=1}^K \sum_{\tau=t_k}^{t_{k+1}-1} c(s_\tau, a_\tau) - \sum_{k=1}^K \tilde{V}_{\ell(t_k)}(s_{init}).
\end{equation}
The algorithm executes the greedy policy $a_\tau = \arg\min_{a \in \mathcal{A}} \left[ c(s_\tau, a) + \min_j \tilde{p}_{\ell(\tau)}^j \tilde{V}_{\ell(\tau)} \right]$, so the chosen action minimizes this expression and $c(s_\tau, a_\tau) = \tilde{V}_{\ell(\tau)}(s_\tau) - \min_j \tilde{p}_{\ell(\tau)}^j \tilde{V}_{\ell(\tau)}$.

To bound the telescoping sum over episodes, we first decompose it using the original value function $\tilde{V}_\ell$. The true trajectory terminates at the goal state, where $\tilde{V}_\ell(g) = 0$:
\begin{align}
    \sum_{k=1}^K \sum_{\tau=t_k}^{t_{k+1}-1} c(s_\tau, a_\tau)
    &= \sum_{\ell=1}^{L_M} \sum_{\tau=t_\ell}^{t_{\ell+1}-1} \Big( \tilde{V}_\ell(s_\tau) - \min_j \tilde{p}_\ell^j \tilde{V}_\ell \Big) \notag \\*
    &= \sum_{\ell=1}^{L_M} \sum_{\tau=t_\ell}^{t_{\ell+1}-1} \underbrace{\Big( \tilde{V}_\ell(s_\tau) - \tilde{V}_\ell(s_{\tau+1}) \Big)}_{\text{(I)}} + \sum_{\tau=1}^{T_K} \Big( \tilde{V}_{\ell(\tau)}(s_{\tau+1}) - \min_j \tilde{p}_{\ell(\tau)}^j \tilde{V}_{\ell(\tau)} \Big).
\end{align}
Because $\tilde{V}_\ell(g) = 0$, term (I) telescopes within each episode and vanishes at its end. Value function jumps only occur at epoch switching moments. The sum of term (I) is therefore:
\begin{equation}
    \sum_{k=1}^K \tilde{V}_{\ell(t_k)}(s_{init}) + \underbrace{ \sum_{\ell=1}^{L_M-1} \Big( \tilde{V}_{\ell+1}(s_{t_{\ell+1}}) - \tilde{V}_\ell(s_{t_{\ell+1}}) \Big) }_{R_{trunc}}.
\end{equation}
Substituting this back into the expression for $R_K$, the initial-state terms cancel with the baseline $-\sum_{k=1}^K \tilde{V}_{\ell(t_k)}(s_{init})$.

Finally, applying the algebraic invariance from Remark \ref{remark:shifted_variable}, we can shift the residual estimation error term to $U_\ell$ without changing its mathematical value:
\begin{equation}
    \tilde{V}_{\ell(\tau)}(s_{\tau+1}) - \min_j \tilde{p}_{\ell(\tau)}^j \tilde{V}_{\ell(\tau)} = U_{\ell(\tau)}(s_{\tau+1}) - \min_j \tilde{p}_{\ell(\tau)}^j U_{\ell(\tau)}.
\end{equation}
By Lemma~\ref{lem:number_epochs}, the total number of epochs satisfies \(L_M=\tilde{\mathcal O}(\sqrt{SAK}+SA)\). Moreover, by Lemma~\ref{lem:value_lower_bound}, the optimistic value function is uniformly bounded in \([-2B_\star,B_\star]\), and hence the value difference incurred at a single epoch switch is at most \(3B_\star\). Therefore, \(R_{\mathrm{trunc}} \le 3B_\star L_M=\tilde{\mathcal O}(B_\star\sqrt{SAK}+B_\star SA)\).
\end{proof}

\begin{lemma}[Extraction of Martingale Difference Sequence]
\label{lem:mds_extract}
By applying Freedman's inequality on the shifted variable $U_\ell$, the cumulative error satisfies with high probability:
\begin{equation}
\begin{split}
    R_K &\le \sum_{\tau=1}^{T_K} \underbrace{\Big( p(\cdot|s_\tau, a_\tau) U_{\ell(\tau)} - \min_{j} \tilde{p}_{\ell(\tau)}^j U_{\ell(\tau)} \Big)}_{\text{Error}_\tau} \\
    &\quad + \Otilde\left( \sqrt{ \sum_{\tau=1}^{T_K} \Var_p(U_{\ell(\tau)})(s_\tau, a_\tau) } + B_\star \right) + \tilde{\mathcal O}(B_\star\sqrt{SAK}+B_\star SA).
\end{split}
\end{equation}
\end{lemma}
\begin{proof}
Adding and subtracting the true single-step expected transition $p(\cdot|s_\tau, a_\tau) U_{\ell(\tau)}$, and substituting the truncation error bound from Lemma \ref{lem:regret_decomp}, we obtain:
\begin{equation}
    R_K \le \sum_{\tau=1}^{T_K} \underbrace{\Big( U_{\ell(\tau)}(s_{\tau+1}) - p(\cdot|s_\tau, a_\tau) U_{\ell(\tau)} \Big)}_{\widetilde{M}_\tau} + \sum_{\tau=1}^{T_K} \text{Error}_\tau + \tilde{\mathcal O}(B_\star\sqrt{SAK}+B_\star SA)
\end{equation}
where $\widetilde{M}_\tau$ is a martingale difference sequence (MDS) with zero conditional expectation. Because $U_\ell \in [0, 3B_\star]$ (Lemma \ref{lem:value_lower_bound}), its absolute value is bounded by $|\widetilde{M}_\tau| \le 3B_\star$. Applying the anytime variance-aware Freedman's inequality (\textbf{Lemma \ref{lem:anytime_freedman}}) with amplitude $3B_\star$, with high probability we have:
\begin{equation}
    \sum_{\tau=1}^{T_K} \widetilde{M}_\tau \le \Otilde\left( \sqrt{ \sum_{\tau=1}^{T_K} \Var_p(U_{\ell(\tau)})(s_\tau, a_\tau) \cdot \log(1/\delta) } + 3B_\star \log(1/\delta) \right) \label{eq:freedman_M}.
\end{equation}
The constant factor $3$ in the residual logarithmic term is absorbed into the $\Otilde$ notation.
\end{proof}

\begin{lemma}[Decomposition of Single-Step Estimation Error]
\label{lem:error_decomp}
The absolute value of the single-step error $\text{Error}_t$ on the shifted variable $U_\ell$ can be decomposed and bounded as:
\begin{equation}
\begin{aligned}
    |\text{Error}_t| \le {}& \underbrace{2\sqrt{\frac{\Var_p(U_\ell) \beta^{KL}}{n_t}} + \frac{\Otilde(B_\star \beta^{KL})}{n_t}}_{(A)}
    + \underbrace{\mathcal{O}\left( \frac{n_{0,t_\ell} B_\star}{n_t} \right)}_{(B)} \\
    &+ \underbrace{2\sqrt{2\kappa_{t_\ell}} \sqrt{\frac{\Var_{\bar{p}_{t_\ell}}(U_\ell) L_\delta}{n_t}} + \frac{\Otilde(\kappa_{t_\ell} B_\star)}{n_t}}_{(C)}.
\end{aligned}
\end{equation}
\end{lemma}
\begin{proof}
Let $t_\ell$ denote the starting step of epoch $\ell$. According to the algorithm's visit-count doubling criterion, within the same epoch, the visit count of any true state-action pair satisfies $n_{t_\ell}(s_t, a_t) \le n_t(s_t, a_t) < 2n_{t_\ell}(s_t, a_t)$, which implies $\frac{1}{n_{t_\ell}} \le \frac{2}{n_t}$.

By introducing the empirical distribution $\hat{p}_{t_\ell}$ and the pseudo-empirical distribution $\bar{p}_{t_\ell}$, we decompose the signed single-step error $\text{Error}_t$ as follows (for brevity, we omit the conditional dependence on $(s_t, a_t)$):
\begin{equation}
    \text{Error}_t = p U_\ell - \min_{j \in [J_{t_\ell}]} \tilde{p}_\ell^j U_\ell = (p - \hat{p}_{t_\ell}) U_\ell + (\hat{p}_{t_\ell} - \bar{p}_{t_\ell}) U_\ell + \max_{j \in [J_{t_\ell}]} (\bar{p}_{t_\ell} - \tilde{p}_\ell^j) U_\ell.
\end{equation}
Applying the triangle inequality, we bound the absolute error by splitting it into three absolute components:
\begin{equation}
    |\text{Error}_t| \le \underbrace{ \big| (p - \hat{p}_{t_\ell}) U_\ell \big| }_{(A)} + \underbrace{ \big| (\hat{p}_{t_\ell} - \bar{p}_{t_\ell}) U_\ell \big| }_{(B)} + \underbrace{ \max_{j \in [J_{t_\ell}]} \big| (\bar{p}_{t_\ell} - \tilde{p}_\ell^j) U_\ell \big| }_{(C)}.
\end{equation}

For term (B), the pseudo-empirical distribution $\bar{p}_{t_\ell}$ is defined as $\frac{n_{t_\ell}}{\bar{n}_{t_\ell}} \hat{p}_{t_\ell} + \frac{n_{0, t_\ell}}{\bar{n}_{t_\ell}} \delta_{s_0}$. Because the shifted variable equals zero at the pseudo-state ($U_\ell(s_0) = 0$), the internal bias term algebraically simplifies:
\begin{equation}
    (\hat{p}_{t_\ell} - \bar{p}_{t_\ell}) U_\ell = \hat{p}_{t_\ell} U_\ell - \left( \frac{n_{t_\ell}}{\bar{n}_{t_\ell}} \hat{p}_{t_\ell} U_\ell + \frac{n_{0,t_\ell}}{\bar{n}_{t_\ell}} U_\ell(s_0) \right) = \frac{n_{0,t_\ell}}{\bar{n}_{t_\ell}} \hat{p}_{t_\ell} U_\ell \ge 0.
\end{equation}
Because this term is nonnegative and $U_\ell \le 3B_\star$, its absolute value equals the term itself:
\begin{equation}
    (B) = \left| (\hat{p}_{t_\ell} - \bar{p}_{t_\ell}) U_\ell \right| = \frac{n_{0,t_\ell}}{\bar{n}_{t_\ell}} \hat{p}_{t_\ell} U_\ell \le \frac{n_{0,t_\ell}}{n_t/2} 3B_\star
\end{equation}
which is bounded by $\mathcal{O}(n_{0,t_\ell} B_\star / n_t)$.

For term (A), under the concentration event $\mathcal{E}^{KL}(\delta)$, because $U_\ell \in [0, 3B_\star]$ is nonnegative, we apply our generalized empirical Bernstein inequality (Lemma \ref{lem:empirical_bernstein_kl}) with $b = 3B_\star$. This inequality provides a two-sided deviation bound, which bounds the absolute difference:
\begin{equation}
    (A) = \big| (p - \hat{p}_{t_\ell}) U_\ell \big| \le \sqrt{\frac{2\Var_p(U_\ell) \beta^{KL}}{n_{t_\ell}}} + \frac{2}{3} (3B_\star) \frac{\beta^{KL}}{n_{t_\ell}} \le 2\sqrt{\frac{\Var_p(U_\ell) \beta^{KL}}{n_t}} + \frac{\Otilde(B_\star \beta^{KL})}{n_t}
\end{equation}
where the constants from the doubling criterion and range extension are absorbed into $\Otilde$.

Term (C) is the maximum absolute deviation of the sampled transition kernels $\tilde{p}_\ell^j \sim \text{Dir}(\frac{\bar{n}_{t_\ell}(s, a)}{\kappa_{t_\ell}(s,a)} \bar{p}_{t_\ell})$ from their mean $\bar{p}_{t_\ell}$. Since $U_\ell \in [0, 3B_\star]$, the Dirichlet concentration bounds apply to $U_\ell$. Let $L_\delta = \mathcal{O}(\log(SA t_\ell/\delta))$ denote the logarithmic factor. On the uniform concentration event $\mathcal{E}^{Dir}(\delta)$, the variance-aware absolute deviation bound holds simultaneously for all samples $j \in [J_{t_\ell}]$, and hence also for the maximum deviation:
\begin{equation}
    (C) = \max_{j \in [J_{t_\ell}]} \big| (\bar{p}_{t_\ell} - \tilde{p}_{\ell}^j) U_\ell \big| \le 2\sqrt{\frac{\kappa_{t_\ell} \cdot \Var_{\bar{p}_{t_\ell}}(U_\ell) L_\delta}{\bar{n}_{t_\ell}}} + \frac{9 \kappa_{t_\ell} \cdot B_\star L_\delta}{\bar{n}_{t_\ell}}.
\end{equation}
Considering the total pseudo-count $\bar{n}_{t_\ell} \ge n_{t_\ell}$, and leveraging the doubling criterion $\frac{1}{\bar{n}_{t_\ell}} \le \frac{2}{n_t}$, we rearrange the lower-order terms to yield:
\begin{equation}
    (C) \le 2\sqrt{\frac{\kappa_{t_\ell} \Var_{\bar{p}_{t_\ell}}(U_\ell) L_\delta}{n_t / 2}} + \frac{18 \kappa_{t_\ell} B_\star L_\delta}{n_t} = 2\sqrt{2\kappa_{t_\ell}} \sqrt{\frac{\Var_{\bar{p}_{t_\ell}}(U_\ell) L_\delta}{n_t}} + \frac{\Otilde(\kappa_{t_\ell} B_\star)}{n_t}.
\end{equation}
Thus, the absolute upper bounds for all three core decomposed components are formally established, completing the proof.
\end{proof}

\begin{lemma}[Variance-Sensitive Transformation of Distributions]
\label{lem:variance_substitution}
Under the concentration event $\mathcal{G}^{conc}(\delta)$, the pseudo-empirical variance $\Var_{\bar{p}_{t_\ell}}(U_\ell)$ in term (C) can be replaced by the true variance $\Var_p(U_\ell)$ while preserving the fast convergence rate. Specifically, there exists an absolute constant $C_0 > 0$ such that the term (C) is bounded by:
\begin{equation}
    (C) \le C_0 \sqrt{\kappa_{t_\ell}} \sqrt{\frac{\Var_p(U_\ell) L_\delta}{n_t}} + \Otilde\left( \frac{\kappa_{t_\ell} B_\star}{n_t} \right).
\end{equation}
\end{lemma}
\begin{proof}
To obtain a tight dependency, we explicitly decompose the variance difference. By definition, $\Var_q(U) = q[U^2] - (qU)^2$. For any distribution $\bar{p}$ and $p$, the difference is expanded as:
\begin{equation}
    \Var_{\bar{p}}(U) - \Var_p(U) = (\bar{p} - p)[U^2] - \big( (\bar{p}U)^2 - (pU)^2 \big).
\end{equation}
Taking the absolute value and applying the triangle inequality, along with the fact that $|(\bar{p}U)^2 - (pU)^2| = |\bar{p}U - pU| \cdot |\bar{p}U + pU|$ and $0 \le U \le 3B_\star$ (thus $\bar{p}U + pU \le 6B_\star$), we obtain the refined perturbation inequality:
\begin{equation}
    |\Var_{\bar{p}}(U_\ell) - \Var_p(U_\ell)| \le \big| (\bar{p}_{t_\ell} - p)[U_\ell^2] \big| + 6B_\star \big| (\bar{p}_{t_\ell} - p) U_\ell \big| \label{eq:var_fine_decomp}.
\end{equation}

Next, we bound the two absolute difference terms respectively by decoupling the pseudo-count bias and the empirical concentration error, noting that $\bar{p} - p = (\bar{p} - \hat{p}) + (\hat{p} - p)$.

\textbf{1. Bounding the linear difference $|(\bar{p}_{t_\ell} - p) U_\ell|$:}
For the pseudo-count bias, since $U_\ell(s_0) = 0$ and $0 \le U_\ell \le 3B_\star$, as shown in Lemma \ref{lem:error_decomp}:
\begin{equation}
    \big| (\bar{p}_{t_\ell} - \hat{p}_{t_\ell}) U_\ell \big| \le \Otilde\left( \frac{B_\star}{n_t} \right).
\end{equation}
For the empirical concentration, applying the empirical Bernstein inequality (\textbf{Lemma \ref{lem:empirical_bernstein_kl}}) on $U_\ell$ under event $\mathcal{E}^{KL}(\delta)$ yields:
\begin{equation}
    \big| (\hat{p}_{t_\ell} - p) U_\ell \big| \le 2\sqrt{\frac{\Var_p(U_\ell) \beta^{KL}}{n_{t_\ell}}} + \frac{\Otilde(B_\star \beta^{KL})}{n_{t_\ell}} \le \Otilde\left( \sqrt{\frac{\Var_p(U_\ell)}{n_t}} + \frac{B_\star}{n_t} \right).
\end{equation}
Combining these, the linear difference is bounded by $\Otilde\left( \sqrt{\frac{\Var_p(U_\ell)}{n_t}} + \frac{B_\star}{n_t} \right)$.

\textbf{2. Bounding the quadratic difference $|(\bar{p}_{t_\ell} - p)[U_\ell^2]|$:}
Similarly, for the pseudo-count bias on the squared variable, since $U_\ell^2(s_0) = 0$ and $0 \le U_\ell^2 \le 9B_\star^2$, we have:
\begin{equation}
    \big| (\bar{p}_{t_\ell} - \hat{p}_{t_\ell})[U_\ell^2] \big| \le \frac{n_{0,t_\ell}}{n_t/2} 9B_\star^2 = \Otilde\left( \frac{B_\star^2}{n_t} \right).
\end{equation}
For the empirical concentration on $U_\ell^2$, we apply the empirical Bernstein inequality on the function $f = U_\ell^2 \in [0, 9B_\star^2]$. The variance of the squared variable is bounded above by the variance of the original variable multiplied by the squared maximum range. Since $U_\ell \in [0, 3B_\star]$, we have $\Var_p(U_\ell^2) \le 4(3B_\star)^2 \Var_p(U_\ell) = 36B_\star^2 \Var_p(U_\ell)$. Substituting this into the Bernstein bound gives:
\begin{equation}
    \big| (\hat{p}_{t_\ell} - p)[U_\ell^2] \big| \le \sqrt{\frac{2\Var_p(U_\ell^2)\beta^{KL}}{n_{t_\ell}}} + \frac{\Otilde(B_\star^2 \beta^{KL})}{n_{t_\ell}} \le \Otilde\left( B_\star \sqrt{\frac{\Var_p(U_\ell)}{n_t}} + \frac{B_\star^2}{n_t} \right).
\end{equation}
Let $c_1, c_2$ be the absolute constants absorbed within the $\Otilde$ notation. Combining these, the quadratic difference is bounded by $c_1 B_\star \sqrt{\frac{\Var_p(U_\ell) \beta_t}{n_t}} + c_2 \frac{B_\star^2 \beta_t}{n_t}$, where $\beta_t$ represents the poly-logarithmic factors.

\textbf{3. Variance Substitution via Case Analysis:}
Substituting the bounds for the linear and quadratic differences back into Eq. \eqref{eq:var_fine_decomp}, we obtain a refined variance deviation bound:
\begin{equation}
    \Delta \triangleq \big| \Var_{\bar{p}_{t_\ell}}(U_\ell) - \Var_p(U_\ell) \big| \le c_1 B_\star \sqrt{\frac{\Var_p(U_\ell) \beta_t}{n_t}} + c_2 \frac{B_\star^2 \beta_t}{n_t} \label{eq:delta_var_bound}.
\end{equation}
To evaluate the pseudo-variance term $\sqrt{\Var_{\bar{p}_{t_\ell}}(U_\ell)}$ in term (C), we use $\Var_{\bar{p}_{t_\ell}}(U_\ell) \le \Var_p(U_\ell) + \Delta$ and consider two cases based on the true variance $a \triangleq \Var_p(U_\ell)$.

\textit{Case 1: High Variance Regime.} Assume $a \ge C \frac{B_\star^2 \beta_t}{n_t}$ for an absolute constant $C$. We specifically choose $C$ to be sufficiently large such that $C \ge 16c_1^2$ and $C \ge 4c_2$. This guarantees that both components of the deviation are dominated by a fraction of the true variance:
\begin{equation}
    c_1 B_\star \sqrt{\frac{a \beta_t}{n_t}} \le \frac{1}{4} a, \qquad \text{and} \qquad c_2 \frac{B_\star^2 \beta_t}{n_t} \le \frac{1}{4} a.
\end{equation}
Therefore, we can bound the total deviation $\Delta$:
\begin{equation}
    \Delta \le c_1 B_\star \sqrt{\frac{a \beta_t}{n_t}} + c_2 \frac{B_\star^2 \beta_t}{n_t} \le \frac{1}{4} a + \frac{1}{4} a = \frac{1}{2} a.
\end{equation}
Consequently, $\Var_{\bar{p}_{t_\ell}}(U_\ell) \le a + \frac{1}{2}a = \frac{3}{2}a$. Taking the square root yields:
\begin{equation}
    \sqrt{\Var_{\bar{p}_{t_\ell}}(U_\ell)} \le \sqrt{\frac{3}{2}} \sqrt{\Var_p(U_\ell)} \label{eq:case1_high_var}.
\end{equation}
This indicates that in the high variance regime, the pseudo-variance is bounded by the true variance up to a minor constant factor.

\textit{Case 2: Low Variance Regime.} Assume $a < C \frac{B_\star^2 \beta_t}{n_t}$. In this regime, the true variance is extremely small, and we bound the deviation $\Delta$ by substituting the condition into Eq. \eqref{eq:delta_var_bound}:
\begin{equation}
    \Delta \le c_1 B_\star \sqrt{\frac{C B_\star^2 \beta_t^2}{n_t^2}} + c_2 \frac{B_\star^2 \beta_t}{n_t} = \Otilde\left( \frac{B_\star^2}{n_t} \right).
\end{equation}
Thus, the total pseudo-variance is uniformly bounded by the fast rate: $\Var_{\bar{p}_{t_\ell}}(U_\ell) \le a + \Delta \le \Otilde\left( \frac{B_\star^2}{n_t} \right)$. Taking the square root yields:
\begin{equation}
    \sqrt{\frac{\Var_{\bar{p}_{t_\ell}}(U_\ell) L_\delta}{n_t}} \le \Otilde\left( \sqrt{\frac{B_\star^2}{n_t^2}} \right) = \Otilde\left( \frac{B_\star}{n_t} \right) \label{eq:case2_low_var}.
\end{equation}

Finally, we substitute the bounds from both regimes back into the original term (C) from Lemma \ref{lem:error_decomp}. In Case 1 (Eq. \eqref{eq:case1_high_var}), the leading term yields $2\sqrt{3\kappa_{t_\ell}} \sqrt{\frac{\Var_p(U_\ell) L_\delta}{n_t}}$. In Case 2 (Eq. \eqref{eq:case2_low_var}), substituting the low variance bound yields $2\sqrt{2\kappa_{t_\ell}} \cdot \Otilde\left( \frac{B_\star}{n_t} \right)$. Since the dynamic inflation parameter satisfies $\kappa_{t_\ell} \ge 1$ by definition, we have $\sqrt{\kappa_{t_\ell}} \le \kappa_{t_\ell}$. Thus, this fractional term can be absorbed into the residual:
\begin{equation}
    2\sqrt{2\kappa_{t_\ell}} \cdot \Otilde\left( \frac{B_\star}{n_t} \right) \le \Otilde\left( \frac{\kappa_{t_\ell} B_\star}{n_t} \right).
\end{equation}
Unifying both cases, there exists an absolute constant $C_0$ (e.g., $C_0 = 2\sqrt{3}$) such that term (C) is bounded by:
\begin{equation}
    (C) \le C_0 \sqrt{\kappa_{t_\ell}} \sqrt{\frac{\Var_p(U_\ell) L_\delta}{n_t}} + \Otilde\left( \frac{\kappa_{t_\ell} B_\star}{n_t} \right).
\end{equation}
This explicitly establishes the variance-sensitive substitution, concluding the proof.
\end{proof}

\begin{lemma}[Upper Bound on Total Variance of True Environment]
\label{lem:total_variance}
The total variance of the true environment along the physical trajectory satisfies:
\begin{equation}
\begin{split}
    \sum_{\tau=1}^{T_K} \Var_p(U_{\ell(\tau)}) &\le \Otilde\left( B_\star^2 K + B_\star R_K + B_\star^2 SA \right) \\
    &\quad + \Otilde(B_\star) \sum_{\tau=1}^{T_K} |\text{Error}_\tau|.
\end{split}
\label{eq:variance_with_R}
\end{equation}
\end{lemma}
\begin{proof}
By the definition of variance, $\Var_p(U_\ell) = \mathbb{E}_p[U_\ell^2] - (\mathbb{E}_p[U_\ell])^2$. From the optimistic Bellman equation shifted by $2B_\star$, we have $U_\ell(s_t) = c(s_t, a_t) + \mathbb{E}_p[U_\ell(s_{t+1})] - \text{Error}_t$.
Rearranging this yields $\mathbb{E}_p[U_\ell(s_{t+1})] = U_\ell(s_t) - c(s_t, a_t) + \text{Error}_t$.
Because $U_\ell(s_t) \ge 0$ by Lemma~\ref{lem:value_lower_bound}, squaring both sides and isolating the expectation gives:
\begin{align}
    \big( \mathbb{E}_p[U_{\ell}(s_{t+1})] \big)^2
    &= (U_\ell(s_t) - c(s_t, a_t) + \text{Error}_t)^2 \nonumber \\
    &= U_{\ell}(s_t)^2 - 2 U_{\ell}(s_t) c(s_t, a_t) + 2 U_{\ell}(s_t) \text{Error}_t + (c(s_t, a_t) - \text{Error}_t)^2 \nonumber \\
    &\ge U_{\ell}(s_t)^2 - 2 U_{\ell}(s_t) c(s_t, a_t) - 2 U_{\ell}(s_t) |\text{Error}_t| \nonumber \\
    &\ge U_{\ell}(s_t)^2 - 6 B_\star c(s_t, a_t) - 6 B_\star |\text{Error}_t| \label{eq:square_expansion}
\end{align}
where the first inequality drops the non-negative squared residual, and the second uses the upper bound $U_\ell(s_t) \le 3B_\star$ and $c \ge 0$.

Substituting Eq. \eqref{eq:square_expansion} into the variance definition yields the single-step variance bound:
\begin{equation}
    \Var_p(U_{\ell(t)}) \le \Big( \mathbb{E}_p[U_{\ell(t)}(s_{t+1})^2] - U_{\ell(t)}(s_t)^2 \Big) + 6 B_\star c(s_t, a_t) + 6 B_\star |\text{Error}_t|.
\end{equation}

Summing this over the global time steps $\tau \in [1, T_K]$, and decomposing the first term into a martingale difference sequence $\widetilde{M}'_\tau$ and a temporal difference term $E_{\text{td}, \tau}$:
\begin{equation}
\begin{split}
    \sum_{\tau=1}^{T_K} \Var_p(U_{\ell(\tau)}) &\le \sum_{\tau=1}^{T_K} \underbrace{\Big( \mathbb{E}_p[U_{\ell(\tau)}(s_{\tau+1})^2] - U_{\ell(\tau)}(s_{\tau+1})^2 \Big)}_{\widetilde{M}'_\tau} \\
    &\quad + \sum_{\tau=1}^{T_K} \underbrace{\Big( U_{\ell(\tau)}(s_{\tau+1})^2 - U_{\ell(\tau)}(s_\tau)^2 \Big)}_{E_{\text{td}, \tau}} \\
    &\quad + 6 B_\star \sum_{\tau=1}^{T_K} c(s_\tau, a_\tau) + 6 B_\star \sum_{\tau=1}^{T_K} |\text{Error}_\tau|.
\end{split}
\label{eq:var_global_sum}
\end{equation}

Next, we bound the first three terms on the right-hand side of Eq. \eqref{eq:var_global_sum} respectively:

\textbf{1. Sum of MDS via Variance-Aware Freedman Inequality:}
Consider the sequence $\widetilde{M}'_\tau$ defined as:
\begin{equation*}
    \widetilde{M}'_\tau = \mathbb{E}_p[U_{\ell(\tau)}(s_{\tau+1})^2] - U_{\ell(\tau)}(s_{\tau+1})^2.
\end{equation*}
This sequence has zero conditional expectation by construction. Furthermore, because $U_\ell \in [0, 3B_\star]$, its absolute value is bounded above by $|\widetilde{M}'_\tau| \le (3B_\star)^2 = 9B_\star^2$.
We now apply the anytime variance-aware Freedman inequality (\textbf{Lemma \ref{lem:anytime_freedman}}). To do so, it remains to bound the predictable variance of the MDS. Since
$0 \le U_{\ell(\tau)} \le 3B_\star$, \textbf{Lemma~\ref{lem:variance_lipschitz}} yields
\begin{equation}
    \Var_p(U_{\ell(\tau)}^2) \le 4(3B_\star)^2 \Var_p(U_{\ell(\tau)})
    = 36B_\star^2 \Var_p(U_{\ell(\tau)}).
\end{equation}
Applying Freedman's inequality with high probability yields:
\begin{align}
    \sum_{\tau=1}^{T_K} \widetilde{M}'_\tau &\le \Otilde\left( \sqrt{\sum_{\tau=1}^{T_K} \Var_p(U_{\ell(\tau)}^2)} + 9B_\star^2 \right) \notag \\
    &\le \Otilde\left( \sqrt{36B_\star^2 \sum_{\tau=1}^{T_K} \Var_p(U_{\ell(\tau)})} + B_\star^2 \right) \notag \\
    &= \Otilde\left( B_\star \sqrt{\sum_{\tau=1}^{T_K} \Var_p(U_{\ell(\tau)})} + B_\star^2 \right). \label{eq:var_term_1_freedman}
\end{align}
This substitution bounds the martingale error in terms of the primary variance quantity, avoiding dependence on the unbounded horizon length $T_K$.

\textbf{2. Sum of Temporal Difference Terms (Telescoping):} Suppose the entire interaction process contains $L_M$ epochs. Within each epoch, this term forms a telescoping sum. However, unlike standard telescoping where the terminal state evaluates to zero, here each episode terminates at the goal state where $U_\ell(g)^2 = (2B_\star)^2 = 4B_\star^2$. Thus, non-zero positive residuals occur both at the $K$ episode terminations and at the $L_M$ epoch switching moments:
\begin{equation}
    \sum_{\tau=1}^{T_K} E_{\text{td}, \tau} \le \sum_{k=1}^K U_{\ell(t_{k+1}-1)}(g)^2 + \sum_{\ell=1}^{L_M-1} \Big( U_{\ell+1}(s_{t_{\ell+1}})^2 - U_{\ell}(s_{t_{\ell+1}})^2 \Big) \le 4 K B_\star^2 + 9 L_M B_\star^2.
\end{equation}
By Lemma~\ref{lem:number_epochs}, the total number of epochs satisfies \(L_M=\tilde{\mathcal O}(\sqrt{SAK}+SA)\). Substituting this yields
\begin{equation}
    \sum_{\tau=1}^{T_K} E_{\text{td}, \tau} \le \Otilde(B_\star^2 K + B_\star^2 \sqrt{SAK} + B_\star^2 SA) = \Otilde(B_\star^2 K + B_\star^2 SA). \label{eq:var_term_2}
\end{equation}

\textbf{3. Bounding the Total Cost:} The total-cost identity follows from the definition of the high-probability regret:
\begin{equation}
    \sum_{\tau=1}^{T_K} c(s_\tau, a_\tau) = K V^\star(s_{init}) + R_K \le K B_\star + R_K \label{eq:var_term_3}.
\end{equation}

Under the strictly positive-cost assumption, the same identity also relates the
random physical time to the cumulative cost:
\begin{equation}
    c_{\min} T_K \le \sum_{\tau=1}^{T_K} c(s_\tau,a_\tau)
    \le K B_\star + R_K^+, \qquad R_K^+ := \max\{R_K,0\}.
    \label{eq:cmin_time_control}
\end{equation}
Consequently, every logarithmic factor involving the random time can be
replaced by
\[
    \log(1+T_K)
    \le \log\!\left(1+\frac{K B_\star+R_K^+}{c_{\min}}\right).
\]
Thus the time-dependent logarithmic factors in the preceding bounds are
controlled by the displayed expression. In the final self-bounding step, the
occurrence of $R_K^+$ inside this logarithm is absorbed, so $c_{\min}$ enters
the regret bound only through logarithmic factors, not through a polynomial
factor in $1/c_{\min}$.

Finally, substituting Eqs. \eqref{eq:var_term_1_freedman}, \eqref{eq:var_term_2}, and \eqref{eq:var_term_3} uniformly back into the global variance summation formula \eqref{eq:var_global_sum}, and letting $X = \sum_{\tau=1}^{T_K} \Var_p(U_{\ell(\tau)})$ be the total variance:
\begin{align}
    X &\le \Otilde\left( B_\star \sqrt{X} + B_\star^2 \right) \notag \\
    &\quad + \Otilde(B_\star^2 K + B_\star^2 SA) \notag \\
    &\quad + 6B_\star(K B_\star + R_K) \notag \\
    &\quad + 6B_\star \sum_{\tau=1}^{T_K} |\text{Error}_\tau|.
\end{align}
To decouple the recursive $\Otilde(B_\star \sqrt{X})$ term, we explicitly apply the AM-GM inequality $c\sqrt{X} \le \frac{1}{2}X + \frac{1}{2}c^2$ with $c = \Otilde(B_\star)$:
\begin{equation}
    \Otilde(B_\star \sqrt{X}) \le \frac{1}{2} X + \Otilde(B_\star^2).
\end{equation}

Moving the term $\frac{1}{2}X$ to the left-hand side gives$
\frac{1}{2}X \le \dots$
and multiplying both sides by $2$ preserves the same $\Otilde(\cdot)$ order.The newly generated constant term $\Otilde(B_\star^2)$ is absorbed by the existing state-complexity term $\Otilde(B_\star^2 SA)$, since $S,A \ge 1$.Moreover, the term $\Otilde(B_\star^2 K)$ is of the same order as the cost contribution $6B_\star^2 K$, so the two can be merged.
This yields the final tightly coupled variance bound:
\begin{equation}
    X \le \Otilde\left( B_\star^2 K + B_\star R_K + B_\star^2 SA \right) + \Otilde(B_\star) \sum_{\tau=1}^{T_K} |\text{Error}_\tau|
\end{equation}
which concludes the proof.
\end{proof}

{
\renewcommand{\thetheorem}{1}
\begin{theorem}[Minimax Regret Bound under Strictly Positive Cost]
\label{thm:main_regret_strict_app}
With probability at least $1-\delta$, under the strictly positive cost assumption ($c_{\min} > 0$), the regret bound of the OPSRL-SSP algorithm satisfies:
\begin{equation}
    R_K \le \Otilde\left( B_\star \sqrt{SAK} + B_\star S^2 A \right) \label{eq:regret_strict_cmin}.
\end{equation}
\end{theorem}
}
\begin{proof}
The proof proceeds in three steps, employing a double self-bounding technique to decouple the error and variance.

\textbf{Step 1: Preliminary bounding of single-step error and regret.}
Combining Lemmas \ref{lem:error_decomp} and \ref{lem:variance_substitution}, we accumulate the single-step errors $\text{Error}_\tau$. Lemma \ref{lem:variance_substitution} bounds the variance transformation without changing the $\Otilde(1/n_t)$ convergence rate. Using the pigeonhole principle and integral relaxation, we bound the sum of these logarithmic decay terms, whose denominators are $n_\tau$, by $\Otilde(B_\star S^2 A)$. We denote this lower-order term by $\text{LOT}_0 = \Otilde( B_\star S^2 A )$.

For the dominant variance-dependent terms, applying the Cauchy-Schwarz inequality yields a preliminary upper bound for the error summation:
\begin{equation}
    \sum_{\tau=1}^{T_K} |\text{Error}_\tau| \le \Otilde(1) \sqrt{ S A \sum_{\tau=1}^{T_K} \Var_p(U_{\ell(\tau)}) } + \text{LOT}_0 \label{eq:error_sum_bound}.
\end{equation}
Substituting Eq. \eqref{eq:error_sum_bound} into Lemma \ref{lem:mds_extract}, and using the truncation bound \(R_{\mathrm{trunc}}=\tilde{\mathcal O}(B_\star\sqrt{SAK}+B_\star SA)\), the regret upper bound can be expressed as
\begin{equation}
    R_K \le \Otilde(1) \sqrt{ S A \sum_{\tau=1}^{T_K} \Var_p(U_{\ell(\tau)}) } + \text{LOT}_1 \label{eq:regret_mid_step}
\end{equation}
where \(\text{LOT}_1=\text{LOT}_0+\tilde{\mathcal O}(B_\star\sqrt{SAK}+B_\star SA)=\tilde{\mathcal O}(B_\star\sqrt{SAK}+B_\star S^2A)\). Here we used \(\text{LOT}_0=\tilde{\mathcal O}(B_\star S^2A)\) and the fact that \(SA\le S^2A\) for \(S\ge 1\).

\textbf{Step 2: Self-bounding decoupling of the total variance term.}
Let $X = \sum_{\tau=1}^{T_K} \Var_p(U_{\ell(\tau)})$. The variance bound given by Lemma \ref{lem:total_variance} incorporates the error summation term $\Otilde(B_\star) \sum |\text{Error}_\tau|$. Substituting the Cauchy-Schwarz upper bound of $\sum |\text{Error}_\tau|$ from Step 1, we obtain an inequality with respect to $X$:
\begin{equation}
    X \le \Otilde\Big( B_\star^2 K + B_\star R_K + B_\star^2 SA \Big) + \Otilde(B_\star \sqrt{SA} \sqrt{X}) + \text{LOT}_2.
\end{equation}
Here, the residual component is $\text{LOT}_2 = \Otilde(B_\star) \text{LOT}_0 = \Otilde( B_\star^2 S^2 A )$.

To decouple $\sqrt{X}$ on the right side, we apply the basic AM-GM inequality $c\sqrt{X} \le \frac{1}{2}X + \frac{1}{2}c^2$ (where $c = \Otilde(B_\star \sqrt{SA})$), relaxing it into the sum of a linear term and a constant term:
\begin{equation}
    \Otilde(B_\star \sqrt{SA} \sqrt{X}) \le \frac{1}{2}X + \Otilde(B_\star^2 SA).
\end{equation}
Substituting this back into the variance inequality and moving $\frac{1}{2}X$ to the left-hand side yields $\frac{1}{2}X \le \Otilde(\dots) + \text{LOT}_2$. Multiplying both sides by 2 removes the recursive dependence on the error term. Combining all terms explicitly, the decoupled total variance bound becomes:
\begin{equation}
    X \le \Otilde\Big( B_\star^2 K + B_\star R_K + B_\star^2 S^2 A \Big)
\label{eq:clean_variance}.
\end{equation}

\textbf{Step 3: Final resolution of the regret bound.}
Substitute the decoupled total variance bound \eqref{eq:clean_variance} back into the regret formulation \eqref{eq:regret_mid_step} from Step 1. Using the basic inequality $\sqrt{\sum x_i} \le \sum \sqrt{x_i}$, we expand the large square root term by term:
\begin{align}
    R_K &\le \Otilde(1) \sqrt{ S A \cdot \Otilde\left( B_\star^2 K + B_\star R_K + B_\star^2 S^2 A \right) } + \text{LOT}_1 \notag \\
    &\le \Otilde\Big( B_\star \sqrt{SAK} + \sqrt{B_\star SA}\sqrt{R_K} + B_\star S^{1.5} A \Big) + \text{LOT}_1.
\end{align}
We now simplify the lower-order terms using basic parameter relationships. Since $S \ge 1$ and $A \ge 1$, we have $B_\star S^{1.5} A \le B_\star S^2 A$. Furthermore, the components of $\text{LOT}_1$ can be absorbed: $B_\star \sqrt{SAK}$ already matches the dominant term, and $B_\star SA \le B_\star S^2 A$.

Grouping these coefficients yields a simplified, standard algebraic quadratic form:
\begin{equation}
    R_K \le a + b\sqrt{R_K}, \label{eq:final_self_bound}
\end{equation}
where the grouped coefficients correspond to:
\begin{align*}
    a &= \Otilde\left( B_\star \sqrt{SAK} + B_\star S^2 A \right), \\
    b &= \Otilde\left( \sqrt{B_\star SA} \right).
\end{align*}
Solving this quadratic self-bounding inequality by completing the square gives:
\begin{align}
    \left(\sqrt{R_K} - \frac{b}{2}\right)^2 \le a + \frac{b^2}{4} \implies R_K \le 2a + 2b^2.
\end{align}
We explicitly evaluate the newly generated component $2b^2$:
\begin{equation}
    2b^2 = \Otilde(B_\star SA).
\end{equation}
Since $2b^2 = \Otilde(B_\star SA)$ is absorbed by the existing state-complexity term $B_\star S^2 A$ in $a$ (given $S \ge 1$), it does not appear in the final bound. Substituting these values back into the expression for $R_K$ yields the final tight regret bound:
\begin{equation}
    R_K \le \Otilde\left( B_\star \sqrt{SAK} + B_\star S^2 A \right).
\end{equation}
This establishes the minimax-optimal leading term while matching the state-of-the-art lower-order complexity, completing the proof.
\end{proof}

\clearpage
\section{Cost Perturbation for Zero-Cost SSPs}
\label{sec:appendix_c_cost_perturbation}

The theoretical bound in Theorem~1 is established under the strictly positive cost assumption $c_{\min}>0$. In this appendix, we show how the same analysis can be extended to general zero-cost SSPs through a universal cost perturbation technique. The purpose of this appendix is purely analytical: we study the perturbed version of the base OPSRL-SSP algorithm and quantify how the perturbation transfers back to the original SSP.

\subsection{Universal Zero-Cost Perturbation}

To handle the general case where zero-cost transitions may be present, we perturb the cost function uniformly:
\begin{equation}
    c_\epsilon(s,a) = \max\{c(s,a), \epsilon\}, \qquad \forall (s,a)\in\mathcal S\times\mathcal A,
\end{equation}
where $\epsilon>0$ is a user-specified perturbation level. Let $\mathcal M_\epsilon$ denote the perturbed SSP instance. By construction, the perturbed model has strictly positive minimum cost
\[
c_{\min}^\epsilon = \epsilon > 0.
\]

Let $V^\star$ and $V_\epsilon^\star$ denote the optimal value functions of the original SSP $\mathcal M$ and the perturbed SSP $\mathcal M_\epsilon$, respectively. We also let $T_\star$ denote the maximum expected hitting time of the optimal policy of the original SSP:
\[
T_\star \triangleq \max_{s\in\mathcal S}\mathbb E_s^{\pi^\star}[\tau_g].
\]

The first step is to quantify the perturbation bias between the original and perturbed regret.

\begin{lemma}[Pathwise Perturbation Transfer]
\label{lem:pathwise_perturbation}
For any realized trajectory generated by a learning algorithm, let $R_K$ denote its cumulative regret in the original SSP $\mathcal M$, and let $R_K^\epsilon$ denote the cumulative regret of the same trajectory evaluated in the perturbed SSP $\mathcal M_\epsilon$. Then, pathwise,
\begin{equation}
    R_K \le R_K^\epsilon + K\epsilon T_\star.
\end{equation}
\end{lemma}

\begin{proof}
Since $c(s,a)\le c_\epsilon(s,a)\le c(s,a)+\epsilon$ for every $(s,a)$, along any realized trajectory we have
\[
\sum_{\tau=1}^{T_K} c(s_\tau,a_\tau)
\le
\sum_{\tau=1}^{T_K} c_\epsilon(s_\tau,a_\tau).
\]
Next, evaluate the optimal policy $\pi^\star$ of the original SSP inside the perturbed model:
\begin{align}
    V_\epsilon^\star(s_{init})
    &\le V_\epsilon^{\pi^\star}(s_{init}) \notag\\
    &= \mathbb E_{\pi^\star}\!\left[\sum_{t=1}^{\tau_g} c_\epsilon(s_t,a_t)\,\middle|\,s_1=s_{init}\right] \notag\\
    &\le \mathbb E_{\pi^\star}\!\left[\sum_{t=1}^{\tau_g} (c(s_t,a_t)+\epsilon)\,\middle|\,s_1=s_{init}\right] \notag\\
    &\le V^\star(s_{init}) + \epsilon T_\star.
\end{align}
Therefore,
\[
-\,V^\star(s_{init}) \le -\,V_\epsilon^\star(s_{init}) + \epsilon T_\star.
\]
Substituting the above inequalities into the definition of regret yields
\begin{align}
    R_K
    &= \sum_{\tau=1}^{T_K} c(s_\tau,a_\tau) - K V^\star(s_{init}) \notag\\
    &\le \sum_{\tau=1}^{T_K} c_\epsilon(s_\tau,a_\tau) - K\bigl(V_\epsilon^\star(s_{init})-\epsilon T_\star\bigr) \notag\\
    &= R_K^\epsilon + K\epsilon T_\star.
\end{align}
\end{proof}

\begin{lemma}[Perturbed Optimal Value Inflation]
\label{lem:perturbed_value_inflation}
The optimal value function in the perturbed SSP satisfies
\begin{equation}
    V_\epsilon^\star(s) \le V^\star(s) + \epsilon T_\star,
    \qquad \forall s\in\mathcal S.
\end{equation}
In particular,
\begin{equation}
    B_\star^\epsilon \triangleq \max_{s\in\mathcal S} V_\epsilon^\star(s)
    \le B_\star + \epsilon T_\star.
\end{equation}
\end{lemma}

\begin{proof}
For any state $s\in\mathcal S$, evaluate the optimal policy $\pi^\star$ of the original SSP in the perturbed model:
\begin{align}
    V_\epsilon^\star(s)
    &\le V_\epsilon^{\pi^\star}(s) \notag\\
    &= \mathbb E_s^{\pi^\star}\!\left[\sum_{t=1}^{\tau_g} c_\epsilon(s_t,a_t)\right] \notag\\
    &\le \mathbb E_s^{\pi^\star}\!\left[\sum_{t=1}^{\tau_g} (c(s_t,a_t)+\epsilon)\right] \notag\\
    &\le V^\star(s)+\epsilon T_\star.
\end{align}
Taking the maximum over $s\in\mathcal S$ and using $V^\star(s)\le B_\star$ completes the proof.
\end{proof}

\subsection{Analysis of OPSRL-SSP on the Perturbed SSP}

We now analyze the base OPSRL-SSP algorithm directly on the perturbed SSP $\mathcal M_\epsilon$. Since $c_{\min}^\epsilon=\epsilon>0$, the entire proof of Theorem~1 applies without modification, provided that the algorithm is supplied with any valid upper bound on the perturbed optimal value function.

\begin{corollary}[Perturbed Base Regret Bound]
\label{cor:perturbed_base_bound}
Suppose OPSRL-SSP (Algorithm~\ref{alg:opsrl_ssp}) is executed on the perturbed SSP $\mathcal M_\epsilon$ with any valid upper bound
\[
B_\epsilon \ge B_\star^\epsilon \triangleq \max_{s\in\mathcal S} V_\epsilon^\star(s).
\]
Then, with probability at least $1-\delta$, its cumulative regret in the perturbed model satisfies
\begin{equation}
    R_K^\epsilon \le \tilde{\mathcal O}\!\left(B_\epsilon \sqrt{SAK} + B_\epsilon S^2A\right).
\end{equation}
\end{corollary}

\begin{proof}
This is a direct application of Theorem~1 to the perturbed SSP $\mathcal M_\epsilon$. Indeed, by construction $\mathcal M_\epsilon$ satisfies the strictly positive cost condition $c_{\min}^\epsilon=\epsilon>0$, and $B_\epsilon$ is assumed to be a valid upper bound on the perturbed optimal value function.
\end{proof}

Combining Corollary~\ref{cor:perturbed_base_bound} with Lemma~\ref{lem:pathwise_perturbation}, we immediately obtain a regret bound in the original SSP.

\begin{theorem}[Cost-Perturbed Regret Bound for OPSRL-SSP]
\label{thm:perturbed_opsrl_ssp}
Fix any perturbation level $\epsilon>0$. Suppose OPSRL-SSP (Algorithm~\ref{alg:opsrl_ssp}) is executed on the perturbed SSP $\mathcal M_\epsilon$ with any valid bound $B_\epsilon \ge B_\star^\epsilon$. Then, with probability at least $1-\delta$, its cumulative regret in the original SSP $\mathcal M$ satisfies
\begin{equation}
    R_K
    \le
    \tilde{\mathcal O}\!\left(
    B_\epsilon \sqrt{SAK}
    +
    B_\epsilon S^2A
    \right)
    +
    K\epsilon T_\star.
\end{equation}
\end{theorem}

\begin{proof}
By Corollary~\ref{cor:perturbed_base_bound}, with probability at least $1-\delta$,
\[
R_K^\epsilon \le \tilde{\mathcal O}\!\left(B_\epsilon \sqrt{SAK}+B_\epsilon S^2A\right).
\]
Applying Lemma~\ref{lem:pathwise_perturbation} yields
\[
R_K \le R_K^\epsilon + K\epsilon T_\star
\le
\tilde{\mathcal O}\!\left(B_\epsilon \sqrt{SAK}+B_\epsilon S^2A\right)
+
K\epsilon T_\star.
\]
\end{proof}

Finally, using Lemma~\ref{lem:perturbed_value_inflation}, one may further upper-bound $B_\star^\epsilon$ by $B_\star+\epsilon T_\star$.

\begin{corollary}[Expanded Form Under Perturbed Value Inflation]
\label{cor:expanded_perturbed_bound}
Suppose the perturbed value bound is chosen so that
\[
B_\epsilon \ge B_\star^\epsilon
\qquad\text{and}\qquad
B_\star^\epsilon \le B_\star+\epsilon T_\star.
\]
Then the regret bound in Theorem~\ref{thm:perturbed_opsrl_ssp} implies
\begin{equation}
\begin{split}
    R_K
    \le
    \tilde{\mathcal O}\!\Big(
    B_\star \sqrt{SAK}
    + B_\star S^2A
    + \epsilon T_\star \sqrt{SAK}
    + \epsilon T_\star S^2A
    \Big)
    + K\epsilon T_\star.
\end{split}
\end{equation}
\end{corollary}

\begin{proof}
By Lemma~\ref{lem:perturbed_value_inflation}, we have $B_\star^\epsilon \le B_\star+\epsilon T_\star$. Substituting this relation into Theorem~\ref{thm:perturbed_opsrl_ssp} yields
\begin{align}
    R_K
    &\le
    \tilde{\mathcal O}\!\left(
    (B_\star+\epsilon T_\star)\sqrt{SAK}
    +
    (B_\star+\epsilon T_\star)S^2A
    \right)
    +
    K\epsilon T_\star \notag\\
    &=
    \tilde{\mathcal O}\!\Big(
    B_\star \sqrt{SAK}
    + B_\star S^2A
    + \epsilon T_\star \sqrt{SAK}
    + \epsilon T_\star S^2A
    \Big)
    + K\epsilon T_\star.
\end{align}
\end{proof}

\begin{remark}[Choice of the Perturbation Level]
Let $\Gamma_K=\sqrt{SAK}+S^2A$. Theorem~\ref{thm:perturbed_opsrl_ssp}
holds for every $0<\epsilon\le1$ and every valid input bound
$B_\epsilon\ge\max\{1,B_\star^\epsilon\}$. To display the dependence on
$\epsilon$, suppose a valid $B_\epsilon$ is available at the scale
$B_\star+\epsilon T_\star$. Then
\[
R_K\le\widetilde{\mathcal O}\!\left(
(B_\star+\epsilon T_\star)\Gamma_K\right)+K\epsilon T_\star.
\]
Unlike the cited PSRL-SSP bound~\citep{jafarniajahromi2021onlinelearningstochasticshortest},
our positive-cost bound introduces no inverse-polynomial factor in
$1/\epsilon$; decreasing $\epsilon$ affects its hidden logarithmic factors.
Up to these logarithms, the explicit perturbation contribution is
$\epsilon T_\star(K+\Gamma_K)$. Thus, $\epsilon=1/(K+\Gamma_K)$ is a
convenient sufficient choice that makes this contribution at most $T_\star$.
When $K$ dominates $\Gamma_K$, the simpler choice $\epsilon=1/K$ has the same
order. We do not claim that either choice exactly minimizes the bound after
expanding the logarithmic factors. The choice of $\epsilon$ uses only $K,S,A$;
as in Theorem~\ref{thm:perturbed_opsrl_ssp}, the guarantee still requires a
valid input bound $B_\epsilon$.
\end{remark}

\clearpage
\section{Auxiliary Technical Lemmas}
\label{sec:appendix_d_aux_lemmas}

In this section, we collect standard concentration inequalities and technical lemmas used throughout our regret analysis. To ensure the paper is self-contained, we restate these results from the original literature and unify the mathematical notation to match the context of our work.

\begin{lemma}[KL Concentration via Method of Types, adapted from \citet{jonsson2020planning} and \citet{tiapkin2022optimisticposteriorsamplingreinforcement}, Theorem C.1]
\label{lem:kl_method_of_types}
Let $p \in \Delta_{m-1}$ be a true categorical distribution with $m$ categories, and $\hat{p}_n$ be the empirical distribution constructed from $n$ independent and identically distributed (i.i.d.) samples. For any $\delta \in (0, 1)$, with probability at least $1-\delta$, it holds simultaneously for all $n \in \mathbb{N}_{++}$ that:
\begin{equation}
    n \cdot \text{KL}(\hat{p}_n \| p) \le \log(1/\delta) + (m-1) \log\left(e\left(1 + \frac{n}{m-1}\right)\right).
\end{equation}
\end{lemma}

\begin{lemma}[Concentration of Lower-Tail KL Divergence, adapted from \citet{tiapkin2022optimisticposteriorsamplingreinforcement}, Theorem C.4]
\label{lem:lower_tail_kl_concentration}
Let $p$ be the true transition distribution and $\hat{p}_n$ be its empirical estimate based on $n$ samples. For any fixed non-negative bounded function $V : \mathcal{S}^+ \to [0, b]$ and any $\delta \in (0, 1)$, the following inequality holds simultaneously for all $n \in \mathbb{N}_{++}$ with probability at least $1-\delta$:
\begin{equation}
    n \cdot \mathcal{K}_{\inf}^{\mathrm{lower}}(\hat{p}_n, p V, V) \le \log(1/\delta) + 3 \log(e\pi(1+2n)).
\end{equation}
\end{lemma}
\begin{proof}
The original literature defines the upper-tail divergence $\mathcal{K}_{\inf}(p, u, f) = \inf \{ \text{KL}(p, q) : qf \ge u \}$ for reward maximization. In our SSP setting, we aim to minimize the cost $V \in [0, b]$. We define an auxiliary function $f(s) = b - V(s)$. Since $V(s) \in [0, b]$, we have $f(s) \in [0, b]$. Observe that the constraint $qV \le u$ is strictly equivalent to:
\begin{equation}
    q(b - f) \le u \iff b - qf \le u \iff qf \ge b - u.
\end{equation}
Thus, there exists a symmetric mapping $\mathcal{K}_{\inf}^{\mathrm{lower}}(\hat{p}_n, pV, V) = \mathcal{K}_{\inf}(\hat{p}_n, b - pV, b - V) = \mathcal{K}_{\inf}(\hat{p}_n, pf, f)$. Applying the upper-tail concentration bound of the original theorem directly to the bounded auxiliary function $f$ gives the stated lower-tail bound.
\end{proof}

\begin{lemma}[Gaussian Lower Bound of Dirichlet Distribution, adapted from \citet{tiapkin2022optimisticposteriorsamplingreinforcement}, Theorem D.2]
\label{lem:dirichlet_anti_concentration}
Let $\theta = (\alpha_0+1, \alpha_1, \dots, \alpha_m) \in \mathbb{R}_{++}^{m+1}$ be the parameters of a Dirichlet distribution. Define the mean distribution $\bar{p} \in \Delta_m$ such that $\bar{p}(0) = (\alpha_0+1)/\bar{\alpha}$ and $\bar{p}(l) = \alpha_l/\bar{\alpha}$ for $l \in \{1, \dots, m\}$, where $\bar{\alpha} = (\alpha_0+1) + \sum_{j=1}^m \alpha_j$.
Let $\epsilon \in (0, 1)$.
Assume the conditions $\alpha_0 \ge c_0(\epsilon) + \log_{17/16}(\bar{\alpha})$ and $\bar{\alpha} \ge 2\alpha_0$ are satisfied, where $c_0(\epsilon)$ is a constant depending only on $\epsilon$. For any function $f : \{0, \dots, m\} \to [0, b_0]$ satisfying $f(0) = b_0$ and $f(j) \le b < b_0/2$ for $j \in \{1, \dots, m\}$, then for any transition probability $\tilde{p} \sim \operatorname{Dir}(\theta)$ sampled from the posterior and any target value $\mu \in (\bar{p}f, b_0)$, we have:
\begin{equation}
    \mathbb{P}_{\tilde{p} \sim \operatorname{Dir}(\theta)} [\tilde{p} f \ge \mu] \ge (1-\epsilon) \mathbb{P}_{g \sim \mathcal{N}(0,1)} \left[ g \ge \sqrt{2\bar{\alpha} \mathcal{K}_{\inf}(\bar{p}, \mu, f)} \right].
\end{equation}
\end{lemma}

\begin{lemma}[Empirical Bernstein Inequality via KL Divergence, adapted from \citet{talebi2018variance} and \citet{tiapkin2022optimisticposteriorsamplingreinforcement}, Lemma E.1]
\label{lem:empirical_bernstein_kl}
Let $p$ and $q$ be any two discrete probability distributions. For any non-negative function $f$ bounded in $[0, b]$, we have:
\begin{align}
    pf - qf &\le \sqrt{2 \Var_q(f) \text{KL}(p, q)} + \frac{2}{3} b \text{KL}(p, q), \\
    qf - pf &\le \sqrt{2 \Var_q(f) \text{KL}(p, q)}.
\end{align}
\end{lemma}

\begin{lemma}[Anytime Variance-Aware Freedman Inequality, adapted from \citet{neu2015exploremoreimprovedhighprobability} and \citet{domingues2020regret}]
\label{lem:anytime_freedman}
Let $\{M_t\}_{t=1}^\infty$ be a martingale difference sequence with respect to a filtration $\mathcal{F}_t$, such that $|M_t| \le b$ holds almost surely. Let $V_t = \sum_{\tau=1}^t \mathbb{E}[M_\tau^2 \mid \mathcal{F}_{\tau-1}]$ be the sum of conditional variances. Then for any $\delta \in (0, 1)$, with probability at least $1-\delta$, it holds simultaneously for all $t \ge 1$ that:
\begin{equation}
    \sum_{\tau=1}^t M_\tau \le \sqrt{2 V_t \log\left(\frac{4e(2t+1)}{\delta}\right)} + 3b \log\left(\frac{4e(2t+1)}{\delta}\right).
\end{equation}
\end{lemma}

\begin{lemma}[Variance contraction under Lipschitz transformation, direct consequence of the Efron--Stein identity \citep{dca15e5b-b3f7-3417-8555-955fe36eb045}]
\label{lem:variance_lipschitz}
Let $X$ be a real-valued random variable supported on an interval $I \subset \mathbb{R}$, and let $g:I\to\mathbb{R}$ be an $L$-Lipschitz function, i.e.,
\[
|g(x)-g(y)| \le L |x-y|, \qquad \forall x,y\in I.
\]
Then
\[
\Var(g(X)) \le L^2 \Var(X).
\]
\end{lemma}
\begin{proof}
Let $X'$ be an independent copy of $X$. Using the identity
\[
\Var(X)=\frac12 \E[(X-X')^2],
\qquad
\Var(g(X))=\frac12 \E[(g(X)-g(X'))^2],
\]
and the $L$-Lipschitz property of $g$, we obtain
\[
\Var(g(X))
=
\frac12 \E[(g(X)-g(X'))^2]
\le
\frac12 \E[L^2 (X-X')^2]
=
L^2 \Var(X).
\]
\end{proof}

\clearpage
\section{Experimental Details and Additional Results}
\label{sec:appendix_e_experiments}

This appendix describes the environments, compared methods, practical implementation of OPSRL-SSP, fixed-instance and held-out evaluations, ablation studies, and compute resources.

\subsection{Environment Descriptions}
We evaluate the algorithms on two widely used tabular SSP environments: \textbf{RandomMDP} and \textbf{GridWorld}.

\textbf{RandomMDP.} The RandomMDP environment is a randomly generated SSP consisting of $S = 8$ states and $A = 2$ actions. The terminal goal state is absorbing with zero cost. For all other non-goal states, the instantaneous cost $c(s, a)$ is drawn uniformly at random from $[0, 1]$. The transition probability matrix for each action is generated by sampling each entry uniformly at random from $[0, 1]$ and then normalizing over the next states.

\textbf{GridWorld.} We implement a standard $10 \times 10$ GridWorld environment, yielding $S = 100$ states and $A = 4$ actions (LEFT, RIGHT, UP, DOWN). The agent starts at the top-left corner $(1,1)$, and the goal state is located at the bottom-right corner $(10,10)$. The environment is a unit-cost grid, meaning $c(s, a) = 1$ for all non-terminal transitions and $c(g, a) = 0$. The transitions are stochastic: when the agent chooses a direction, it successfully moves in that intended direction with a probability of $0.7$. With a total probability of $0.3$ (i.e., $0.1$ for each of the other three directions), the agent slides into an unintended adjacent state. If a movement attempts to cross the grid boundary, the agent remains in its current position.

\subsection{Compared Methods}
We compare OPSRL-SSP with UC-SSP \citep{tarbouriech2020noregretexplorationgoalorientedreinforcement}, Bernstein-SSP \citep{cohen2020nearoptimalregretboundsstochastic}, ULCVI \citep{cohen2021minimaxregretstochasticshortest}, EB-SSP \citep{tarbouriech2021stochasticshortestpathminimax}, SVI-SSP, LCB-ADVANTAGE-SSP \citep{chen2021implicitfinitehorizonapproximationefficient}, and PSRL-SSP \citep{jafarniajahromi2021onlinelearningstochasticshortest}. We additionally include $\epsilon$-greedy Q-learning as a heuristic reference in the fixed-instance experiments; the reported held-out comparison focuses on PSRL-SSP and model-based OFU baselines.

\subsection{Practical Implementation of OPSRL-SSP}

While our theoretical analysis in Section 3 requires the dynamic posterior inflation \(\kappa_t\), dynamic pseudo-counts \(n_{0,t}\), and a logarithmically growing number of posterior samples \(J_t\) to guarantee frequentist anytime optimism, using these theoretical parameter settings can be computationally intensive and overly conservative in practice.

The role of these dynamic parameters is primarily theoretical. In particular, the logarithmic growth of \(J_t\) is introduced to make the optimism event hold uniformly over all time steps and all state-action pairs through an anytime union bound. It should not be interpreted as an empirical recommendation to increase exploration indefinitely. Since OPSRL-SSP plans with the most optimistic transition model among \(J_t\) posterior samples, a larger \(J_t\) amplifies lower-tail optimism and can lead to over-exploration in benign stochastic environments. Similarly, the dynamic pseudo-count \(n_{0,t}\) is chosen to guarantee Dirichlet anti-concentration in the worst case, but may be conservative when the posterior has already concentrated around the true dynamics. Therefore, as is common in empirical evaluations of OFU-style algorithms where theoretical confidence radii are scaled down, we use small fixed values of \(J\) and the optimistic pseudo-mass in our implementation.

To bridge this theory-practice gap, our practical implementation of OPSRL-SSP fixes these three mechanisms to constant hyper-parameters:
\begin{enumerate}[leftmargin=*]
    \item \textbf{Number of Posterior Samples ($J$):} Controlled by \texttt{sample\_cap}. We set $J=1$ in both environments. As we will demonstrate in the subsequent ablation study, a single optimistic posterior sample is empirically sufficient to drive exploration, reducing the computational overhead compared to maintaining $J \approx \log(T)$.
    \item \textbf{Base Dirichlet Prior:} Controlled by \texttt{prior}. We use a constant base Dirichlet prior (e.g., $0.0095$) for standard physical transitions, providing a mild initial regularization similar to standard PSRL.
    \item \textbf{Optimistic Pseudo-count Mass:} Controlled by \texttt{pseudo\_mass}. Instead of dynamically scaling $n_{0,t}$ with the empirical variance, we inject a fixed amount of pseudo-count mass to the optimistic pseudo-state $s_0$ at the start of each epoch.
\end{enumerate}
This fixed-parameter approach retains the structural mechanism of the optimistic boundary condition ($\tilde{V}_\ell(s_0) = -2B_\star$) while retaining an efficient exploration mechanism.

\subsection{Fixed-instance RandomMDP and GridWorld Evaluation}

The fixed RandomMDP instance uses seed $6$, for which $B_\star\approx2.31$ and $c_{\min}\approx0.0417$. The fixed GridWorld instance uses seed $1$, for which $B_\star\approx28.66$ and $c_{\min}=1$.

On each fixed instance, every method is evaluated over 500 independent runs of 5,000 episodes. Figure~\ref{fig:plot} shows the mean cumulative regret, with pointwise 95\% confidence intervals across runs.

Following the methodology of recent empirical studies \citep{chen2021implicitfinitehorizonapproximationefficient}, instead of using the exact theoretically derived logarithmic confidence intervals (which often lead to overly conservative and poor empirical performance), we treat the scaling factor of the bonuses as a tunable hyper-parameter $\iota$. For each baseline, we perform a grid search to select the parameters that yield the best empirical regret. The hyper-parameters used in the fixed-instance evaluations are summarized in Table~\ref{tab:hyperparameters}.

\begin{table}[htbp]
\centering
\caption{Hyper-parameters used in the fixed-instance RandomMDP and GridWorld experiments.}
\label{tab:hyperparameters}
\small
\begin{tabular}{@{}lll@{}}
\toprule
\textbf{Algorithm} & \textbf{RandomMDP} & \textbf{GridWorld} \\
\midrule
Q-learning ($\epsilon$-greedy) & $\epsilon = 0.05$ & $\epsilon = 0.05$ \\
LCB-ADVANTAGE-SSP & $H = 5$, $\iota = 0.05$, $\theta^\star = 4096$ & $H = 30$, $\iota = 0.1$, $\theta^\star = 4096$ \\
UC-SSP & $\iota = 1.0$ & $\iota = 1.0$ \\
Bernstein-SSP & $\iota = 2.0$ & $\iota = 2.0$ \\
ULCVI & $H = 80$, $\iota = 2.0$ & $H = 200$, $\iota = 1.0$ \\
EB-SSP & $\iota = 0.05$ & $\iota = 0.01$ \\
SVI-SSP & $H = 15$, $\iota = 0.05$, $B=1.5$ & $H = 50$, $\iota = 0.01$, $B=30.0$ \\
PSRL-SSP & default & default \\
\midrule
\textbf{OPSRL-SSP} & \texttt{pseudo\_mass} = 0.05, & \texttt{pseudo\_mass} = 0.02, \\
& \texttt{sample\_cap} ($J$) = 1, & \texttt{sample\_cap} ($J$) = 1, \\
& \texttt{prior} = 0.0095, $\delta = 0.001$ & \texttt{prior} = 0.0098, $\delta = 0.001$ \\
\bottomrule
\end{tabular}
\end{table}

Figure~\ref{fig:plot} compares learning on two fixed SSP instances with different state spaces and cost structures. In RandomMDP, OPSRL-SSP is among the lower-regret methods: it outperforms Q-learning and LCB-ADVANTAGE-SSP and remains close to PSRL-SSP, EB-SSP, and Bernstein-SSP. In GridWorld, OPSRL-SSP again lies in the lower-regret group, with performance close to PSRL-SSP and EB-SSP and below Q-learning, LCB-ADVANTAGE-SSP, and Bernstein-SSP.

ULCVI and SVI-SSP also perform strongly on these fixed instances. ULCVI uses an explicit finite-horizon reduction with terminal costs, whereas SVI-SSP uses sparse, stage-wise value updates with a Bernstein-style bonus. OPSRL-SSP follows a posterior-sampling route and does not require either of these particular finite-horizon or explicit-bonus constructions. Thus, the fixed-instance curves show that posterior-generated optimism remains empirically competitive across two distinct SSP structures. This empirical observation complements the minimax-optimal leading term established in Theorem~1.

\subsection{Held-out RandomMDP Evaluation}
\label{sec:app_e_heldout_randommdp}

The held-out cohort consists of 20 independently generated RandomMDP instances, with environment seeds $841001,842001,\ldots,860001$. Each instance has the same $S=8$, $A=2$ construction described above; its sampled costs and transition kernel are shared across methods and fixed across their 20 rollouts. Each rollout lasts 5,000 episodes, and the plotted metric is cumulative sample-path regret. At each checkpoint, we first average the 20 rollouts within each instance, then compute the mean and pointwise 95\% Student-$t$ confidence interval from the resulting 20 instance-level curves.

The configurations underlying Figure~\ref{fig:heldout_randommdp} are listed in Table~\ref{tab:heldout_randommdp_configs}. The OPSRL-SSP curve uses the practical RandomMDP parameters reported in Table~\ref{tab:hyperparameters}. SVI-SSP is included as a configuration-specific diagnostic with $B=5.1$, which exceeds the audited maximum $B_\star\approx4.9853$ among these 20 instances.

\begin{table}[htbp]
\centering
\small
\caption{Practical configurations in the held-out RandomMDP figure.}
\label{tab:heldout_randommdp_configs}
\begin{tabular}{@{}ll@{}}
\toprule
\textbf{Method} & \textbf{Configuration} \\
\midrule
OPSRL-SSP & \texttt{pseudo\_mass}$=0.05$, \texttt{prior}$=0.0095$, $J=1$ \\
PSRL-SSP & default \\
ULCVI & $H=80$, $\iota=2.0$ \\
EB-SSP & $\iota=0.05$ \\
LCB-ADVANTAGE-SSP & $H=5$, $\iota=0.05$, $\theta^\star=4096$ \\
UC-SSP & $c=1.0$ \\
Bernstein-SSP & $c=2.0$ \\
SVI-SSP (diagnostic) & $H=20$, $\iota=1.0$, $B=5.1$ \\
\bottomrule
\end{tabular}
\end{table}

Table~\ref{tab:heldout_randommdp_results} reports the endpoint $C(5000)$ and the normalized area under each cumulative-regret curve, computed by trapezoidal integration over the recorded checkpoints and divided by 5,000. Each entry is the mean $\pm$ 95\% Student-$t$ confidence-interval half-width across the 20 instance-level values. On this cohort, OPSRL-SSP has lower endpoint and nAUC point estimates than the tested ULCVI, EB-SSP, LCB-ADVANTAGE-SSP, Bernstein-SSP, and SVI-SSP configurations; PSRL-SSP and UC-SSP have lower point estimates.

\begin{table}[htbp]
\centering
\small
\caption{Held-out RandomMDP endpoint regret and nAUC across 20 independently generated instances. Intervals are environment-level 95\% Student-$t$ half-widths.}
\label{tab:heldout_randommdp_results}
\begin{tabular}{@{}lcc@{}}
\toprule
\textbf{Method} & \textbf{$C(5000)$} & \textbf{nAUC} \\
\midrule
OPSRL-SSP & $252.051\pm46.935$ & $185.038\pm30.329$ \\
PSRL-SSP & $233.178\pm65.217$ & $167.753\pm40.360$ \\
ULCVI & $940.474\pm376.049$ & $492.066\pm188.669$ \\
EB-SSP & $987.534\pm123.123$ & $801.472\pm100.701$ \\
LCB-ADVANTAGE-SSP & $530.668\pm175.800$ & $337.898\pm106.638$ \\
UC-SSP & $143.653\pm59.317$ & $102.867\pm34.301$ \\
Bernstein-SSP & $274.980\pm61.959$ & $194.634\pm39.850$ \\
SVI-SSP ($B=5.1$; diagnostic) & $2152.487\pm163.344$ & $1549.865\pm144.086$ \\
\bottomrule
\end{tabular}
\end{table}

\subsection{Ablation Studies}

To validate our practical fixed-parameter design and examine the mechanisms of OPSRL-SSP, we isolate the effects of the posterior sample count ($J$) and the optimistic prior strength ($n_0$) via two ablation studies in the RandomMDP environment.

\paragraph{Impact of the Number of Posterior Samples ($J$).}
Our theoretical analysis formally requires the number of posterior samples to scale logarithmically, i.e., $J \propto \log(T)$, to establish the uniform optimism guarantee. To evaluate the empirical consequence of fixing this parameter, we kept the prior configuration constant and varied $J \in \{1, 4, 8, 16, 32\}$.

\begin{figure}[htbp]
    \centering
    \includegraphics[width=0.5\textwidth]{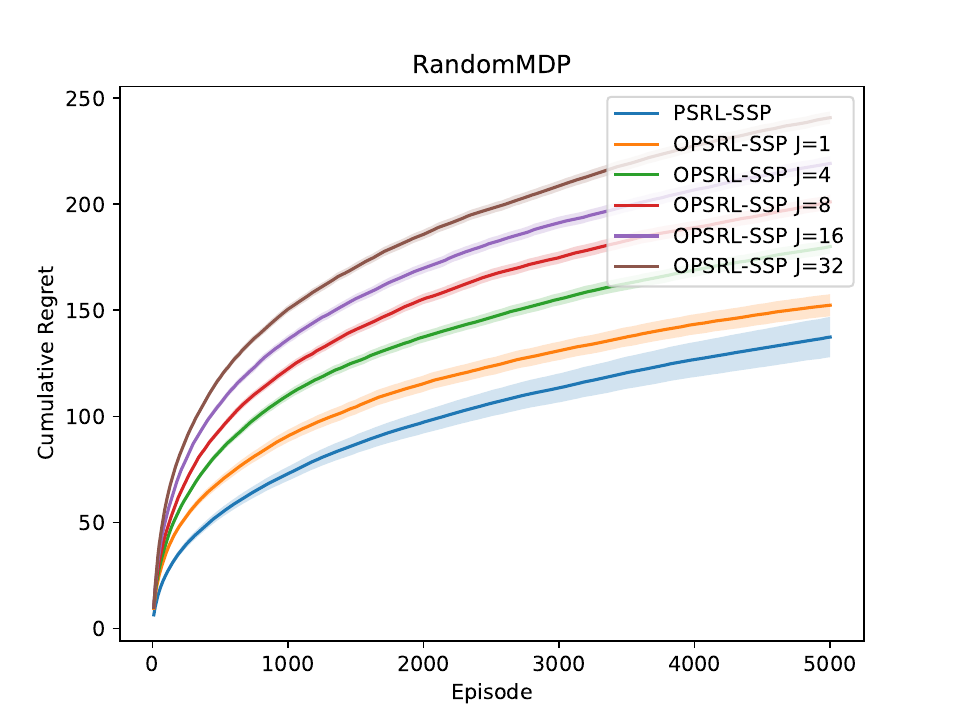}
    \caption{Cumulative regret of OPSRL-SSP for varying numbers of posterior samples $J \in \{1, 4, 8, 16, 32\}$.}
    \label{fig:j_samples}
\end{figure}

As shown in Figure \ref{fig:j_samples}, increasing the number of posterior samples $J$ results in a slight increase in cumulative regret. This observation demonstrates the divergence between worst-case theoretical guarantees and average-case empirical performance. The theoretically prescribed schedule \(J_t \propto \log(T)\) is introduced as a sufficient condition for guaranteeing optimism uniformly over time under worst-case transition dynamics. However, in standard stochastic environments such as RandomMDP, evaluating multiple transition models and planning greedily with respect to the minimum expected cost introduces excessive optimism, which consequently induces over-exploration. This empirical behavior is consistent with the tuning practices in OFU-based algorithms, where theoretically derived confidence radii are systematically scaled down (e.g., setting the hyper-parameter $\iota \ll 1$) to yield computationally efficient and practically competitive policies.

Furthermore, consistent with findings in episodic OPSRL \citep{tiapkin2022optimisticposteriorsamplingreinforcement}, the variation in regret caused by $J$ is marginal compared to the significant performance gap between minimax-optimal algorithms and sub-optimal baselines. Consequently, setting $J=1$ coupled with the pseudo-state mechanism provides sufficient optimistic exploration for average-case scenarios, yielding a compute-efficient configuration without compromising empirical performance.

\paragraph{Impact of the Optimistic Prior Strength ($n_0$).}
In contrast to the monotonic behavior of $J$, varying the prior mass allocation exhibits a structural trade-off. To isolate the effect of the optimistic bias without altering the overall regularization magnitude, we fixed $J=1$ and the \textit{total} initial prior mass at exactly $0.126$. We then varied the allocation of this mass to the optimistic pseudo-state $n_0 \in \{0.00, 0.01, 0.02, 0.05, 0.08, 0.10\}$, distributing the residual mass equally to provide uniform Dirichlet smoothing over the physical transitions.

\begin{figure}[htbp]
    \centering
    \includegraphics[width=0.5\textwidth]{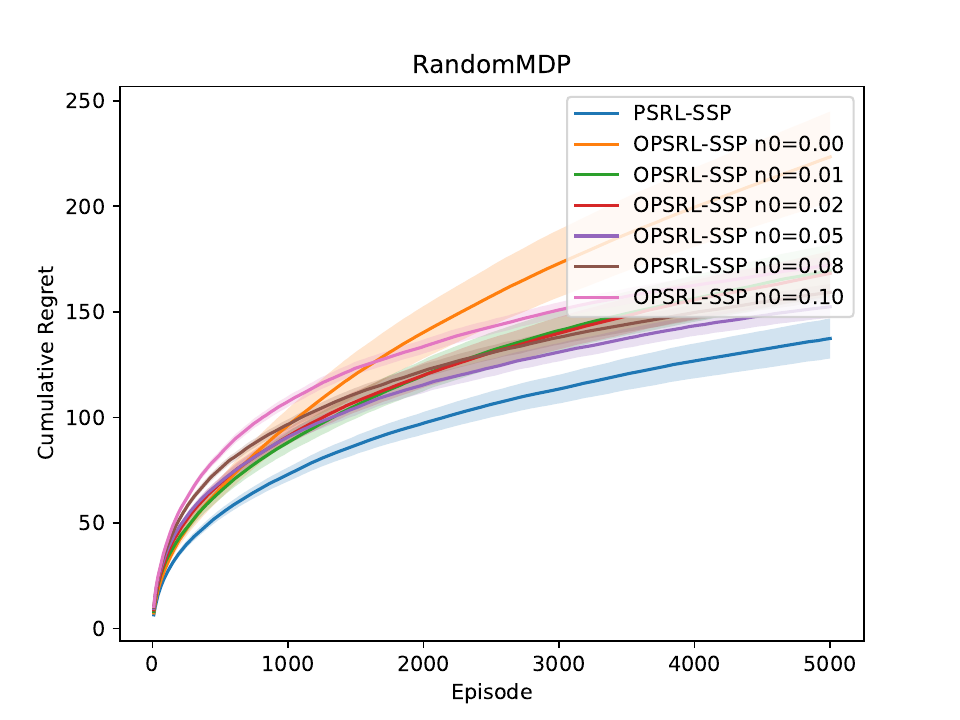}
    \caption{Cumulative regret of OPSRL-SSP under a fixed total prior mass, varying the optimistic pseudo-mass $n_0$.}
    \label{fig:n0_mass}
\end{figure}

As shown in Figure \ref{fig:n0_mass}, the regret exhibits a U-shaped trend. This non-monotonicity arises from a trade-off under a fixed total mass budget: increasing $n_0$ explicitly strengthens the optimistic bias, but concurrently diminishes the uniform Dirichlet smoothing over the real states, which subsequently increases the variance of the posterior updates. When $n_0=0.00$, the absence of optimism results in insufficient exploration. Conversely, when $n_0=0.10$, the optimistic prior becomes disproportionately dominant, requiring a substantially larger amount of empirical data to converge to the true environment dynamics, thereby inducing a persistent sub-optimal bias. The optimal balance is achieved at $n_0=0.05$, which performs comparably to the PSRL-SSP baseline while retaining minimax guarantees.

Together, these results show a difference from standard episodic reward-maximization settings: in our pseudo-state-augmented SSP adaptation, $J$ directly amplifies the extremeness of the statistical optimism, while $n_0$ regulates the bias-variance trade-off of the posterior distribution. Fixing a mild $n_0$ alongside $J=1$ is therefore a theoretically grounded and computationally efficient practical choice.

\subsection{Compute Resources}
All experiments, including the fixed-instance benchmarks and the held-out RandomMDP cohort, were run on a dual-socket Intel Xeon Platinum 8475B server with approximately 251\,GiB of RAM. Although the server also had two NVIDIA RTX 5090 GPUs, the tabular experiments used CPUs only. The empirical comparisons concern learning behavior rather than large-scale systems performance.

\end{document}